\documentclass[10pt,twocolumn,letterpaper]{article}

\usepackage[pagenumbers]{cvpr} 

\definecolor{cvprblue}{rgb}{0.21,0.49,0.74}
\usepackage[pagebackref,breaklinks,colorlinks=true,citecolor=cvprblue]{hyperref}
\usepackage{multirow}
\usepackage{makecell}
\usepackage{pifont}
\usepackage{placeins}
\usepackage[accsupp]{axessibility}

\def\paperID{7239} 
\def\confName{CVPR}
\def\confYear{2025}

\definecolor{darkred}{rgb}{0.7, 0.0, 0.0}
\definecolor{darkred2}{rgb}{0.5, 0.0, 0.0}
\definecolor{darkred3}{rgb}{0.9, 0.0, 0.0}
\definecolor{darkgreen}{rgb}{0.0, 0.42, 0.24}
\definecolor{darkblue}{rgb}{0.10, 0.17, 0.8}
\definecolor{Gray}{gray}{0.93}

\def\eg{\emph{e.g.}} 
\def\ie{\emph{i.e.}} 
\def\etc{\emph{etc.}\xspace}

\newcommand{\method}{DeQA-Score\xspace}
\newcommand{\supp}{Appendix\xspace}

\title{Teaching Large Language Models to Regress Accurate Image Quality Scores Using Score Distribution}

\author{Zhiyuan You$^{12}$, Xin Cai$^2$, Jinjin Gu$^4$, Tianfan Xue$^{235}$$^\dag$, Chao Dong$^{134}$$^\dag$ \vspace{2pt} \\
$^1$Shenzhen Institutes of Advanced Technology, Chinese Academy of Sciences \\
$^2$Multimedia Laboratory, The Chinese University of Hong Kong, $^3$Shanghai AI Laboratory \\
$^4$Shenzhen University of Advanced Technology, $^5$CPII under InnoHK \\
{\tt\small zhiyuanyou@link.cuhk.edu.hk, caixin@link.cuhk.edu.hk, jinjin.gu@sydney.edu.au} \\
{\tt\small tfxue@ie.cuhk.edu.hk, chao.dong@siat.ac.cn} \quad 
{\small $^\dag$ Corresponding Author} \\
}

\begin{document}
\maketitle
\vspace{-10pt}
\begin{abstract}

With the rapid advancement of Multi-modal Large Language Models (MLLMs), MLLM-based Image Quality Assessment (IQA) methods have shown promising performance in linguistic quality description.
However, current methods still fall short in accurately scoring image quality. 
In this work, we aim to leverage MLLMs to regress accurate quality scores. 
A key challenge is that the quality score is inherently \textbf{continuous}, typically modeled as a Gaussian distribution, whereas MLLMs generate \textbf{discrete} token outputs. 
This mismatch necessitates score discretization. 
Previous approaches discretize the mean score into a one-hot label, resulting in information loss and failing to capture inter-image relationships. 
We propose a distribution-based approach that discretizes the score distribution into a soft label. 
This method preserves the characteristics of the score distribution, achieving high accuracy and maintaining inter-image relationships. 
Moreover, to address dataset variation, where different IQA datasets exhibit various distributions, we introduce a fidelity loss based on Thurstone's model. 
This loss captures intra-dataset relationships, facilitating co-training across multiple IQA datasets. 
With these designs, we develop the distribution-based \textbf{De}picted image \textbf{Q}uality \textbf{A}ssessment model for \textbf{Score} regression (\method).
Experiments across multiple benchmarks show that \method stably outperforms baselines in score regression. 
Also, \method can predict the score distribution that closely aligns with human annotations. 
Codes and model weights have been released in \url{https://depictqa.github.io/deqa-score/}.

\end{abstract}
\vspace{-10pt}

\section{Introduction}\label{sec:intro}

\begin{figure}[t]
    \centering
    \includegraphics[width=0.95\linewidth]{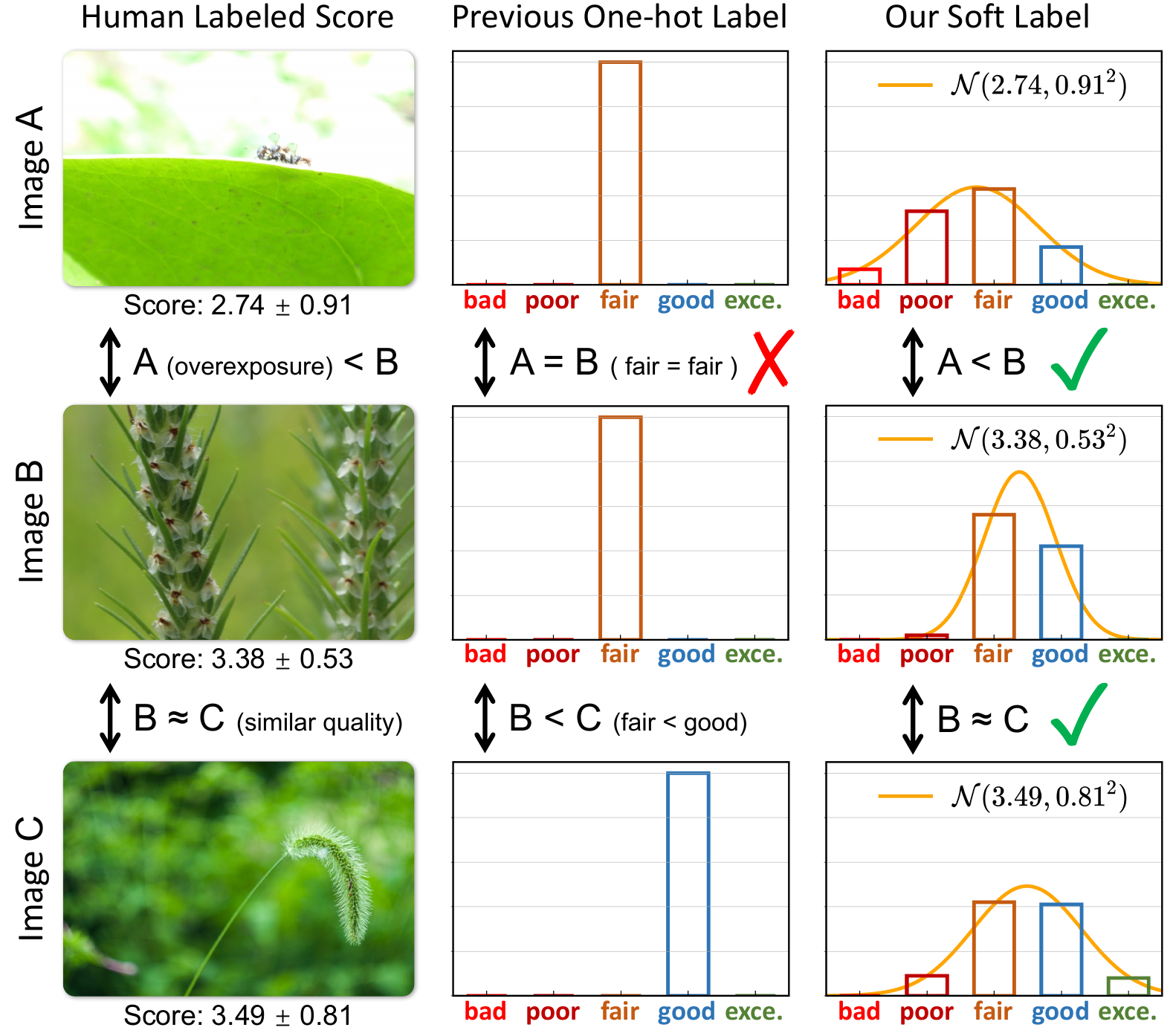}
    \vspace{-5pt}
    \caption{
    \textbf{Illustration of our distribution-based soft label}.
    Image A is overexposed, while Image B and C are well-exposed with better and similar quality. 
    ``exce.'' means ``excellent''. 
    To train an MLLM as a quality scorer, continuous scores need to be discretized to discrete level tokens as the training label. 
    Q-Align~\cite{qalign} discretizes human-labeled mean score to a one-hot label, causing information loss. 
    We discretize the score distribution, which is approximated as a Gaussian distribution, to obtain a more accurate soft label. 
    Our soft label maintains better inter-image relationships that the Image B is better than A (${\rm A < B}$), and Image B and C share similar quality (${\rm B \approx C}$), while the one-hot label can not. 
    }
    \label{fig:teaser}
    \vspace{-15pt}
\end{figure}


Image Quality Assessment (IQA) focuses on perceiving the visual quality of images~\cite{lpips, musiq, ssim, niqe}, and is widely adopted in various application scenarios, \eg, image compression and transmission~\cite{live, phocolens}, processing smartphone-captured photos~\cite{spaq}, and evaluating generation or restoration models~\cite{agiqa, agenticir, ultrafusion, prompt_dehaze}, \etc 
With the emergence of Multi-modal Large Language Models (MLLMs)~\cite{llava, mplugowl2, minigpt4, gpt4v}, MLLM-based IQA begins to attract more research interests~\cite{qinstruct, coinstruct, qground, depictqa, depictqav2}. 
These methods mainly leverage MLLMs to describe image quality using
language, recognizing that language better mirrors human expression, and captures the multifaceted nature of IQA tasks~\cite{depictqa}.

However, for quality scoring, which needs an accurate numerical score to quantify an image's quality, existing description-centered MLLMs~\cite{qinstruct, coinstruct, depictqa, depictqav2} perform less accurately than traditional IQA methods. 
This hinders their applications on real-world IQA scenarios. 
Considering MLLMs' good generalization ability on various downstream tasks and promising performance on quality description, it is possible for MLLMs to score the image quality. 
Therefore, in this work, we aim to explore \textit{how MLLMs can effectively regress accurate quality scores} by leveraging their broad training and inherent adaptability.


We find that the principal challenge in training MLLMs to predict accurate quality scores is the distribution gap between MLLMs' outputs and quality scores. 
MLLMs generate outputs as \textit{discrete} tokens, while the quality score is inherently \textit{continuous}, usually approximated to a Gaussian distribution. 
Previous approaches, like Q-Align~\cite{qalign}, attempte to address the distribution gap by directly discretizing the mean score (\ie, the expectation of estimated score distribution) into a one-hot label, as depicted in \cref{fig:teaser}. 
However, such a one-hot discretization method introduces information loss, causing mismatches between the discretized labels and the original continuous scores. 
For instance, in \cref{fig:teaser}, Image A is overexposed with a lower mean score than Image B, but their one-hot labels appear identical. 
Conversely, Image B and C are well-exposed with similar quality scores, yet their discretized labels are different.

To overcome the distribution gap, we propose a novel soft label to bridge continuous and discrete distributions. 
As illustrated in \cref{fig:teaser}, instead of discretizing the mean of score distribution, we discretize the continuous score distribution itself into a discrete distribution of different levels. 
During training, this discrete distribution is used as the soft training label. 
This enables effective training with minimal information loss. 
Also, our distribution-based soft label better preserves inter-image relationship. 
For example, in \cref{fig:teaser}, Image B is better than Image A, and Image B and C share similar quality, while the simple one-hot label fails to capture such relationships. 
At inference, MLLMs output discrete token results, which can be recovered to continuous score distribution. 
This method ensures that the discrete outputs align well with the underlying continuous distribution, allowing to accurately predict the desired score.


Another key challenge is the dataset variation, where different IQA datasets exhibit distinct distributions, making it hard to simply combine them for co-training. 
Due to this variation, images from different datasets may have nearly the same quality scores (\ie, linearly re-scaled scores) but drastically different perceptual quality~\cite{unique, compare2score}. 
A previous method, UNIQUE~\cite{unique}, has demonstrated that the fidelity loss~\cite{fidelity} under the Thurstone’s model~\cite{thurstone} can effectively capture the intra-dataset relationships, thus facilitating the joint training of multiple IQA datasets. 
However, the fidelity loss requires the model to predict an accurate score distribution, which MLLMs trained with the one-hot label cannot provide. 
Fortunately, MLLMs trained with our distribution-based soft label can predict such an accurate score distribution, allowing the adaptation of the fidelity loss. 
We apply this fidelity loss to our MLLM-based quality scorer and demonstrate its effectiveness in this new context.


Equipped with these designs, we successfully develop a distribution-based multi-modal image quality assessment model, termed \method. 
\method achieves better score regression results than baseline methods on both in-distribution (\eg, 1.3\% in PLCC on KonIQ~\cite{koniq}) and out-of-distribution datasets (\eg, 4.6\% in PLCC on LIVE-WILD~\cite{livewild}). 
The advantage of our \method becomes more pronounced with the co-training across multiple IQA datasets. 
Moreover, our method can further predict the distribution of quality scores, which is closely aligned with the distribution derived from extensive human annotations. 
We hope that our \method will inspire more research to explore the potential of quality score distribution.

\section{Related Works}
\label{sec:related_works}

\textbf{Image Quality Assessment} (IQA) methods mainly rely on quality scores to assess image quality. 
They can be categorized into \textit{full-reference} and \textit{non-reference} methods through whether there is a high-quality reference image.

Full-reference IQA methods compute a similarity score as the quality score between a distorted image and a high-quality reference. 
Classical works rely on human-designed metrics such as structural similarity~\cite{ssim}, phase congruency with gradient magnitude~\cite{fsim}, image information~\cite{vif}, \etc
The rapid advancement of deep learning has also inspired learning-based IQA methods that measure image quality through data-driven training. 
Pioneered by LPIPS~\cite{lpips} and PieAPP~\cite{pieapp}, driven by large-scale IQA datasets~\cite{lpips, pipal, kadid, live}, deep-learning methods~\cite{WaDIQaM, JSPL, dists, A-DISTS, ghildyal2022stlpips, CVRKD, SRIF, afine} have exhibited quite high accuracy in quality score regression.

Non-reference IQA methods directly regress a quality score without a reference image. 
Initially, hand-crafted natural image statistics are adopted~\cite{ma2017learning, brisque, niqe, moorthy2010two, DIIVINE, saad2012blind, tang2011learning}. 
Subsequently, deep-learning-based methods~\cite{CNNIQA, RankIQA, BPSQM, HyperIQA, graphiqa, CKDN, MetaIQA, zhang2022continual, paq2piq} replace hand-crafted statistics by learning quality priors from human-annotated IQA datasets~\cite{koniq, spaq, kadid, agiqa}. 
Recent works further improve the regression accuracy by introducing multi-dataset co-training strategy~\cite{unique}, multi-scale features~\cite{musiq}, CLIP pre-training~\cite{clipiqa}, multi-dimension attention~\cite{maniqa}, multitask learning~\cite{liqe}, and so on.

\vspace{2pt}\noindent\textbf{Multi-modal Large Language Models} (MLLMs) incorporate visual modality into Large Language Models (LLMs)~\cite{vicuna, gpt4, llama, internlm} to achieve general visual ability by leveraging the broad training and powerful generalization ability of LLMs.  
Existing MLLMs~\cite{flamingo, instructblip, llava, gpt4v, vary, mplugowl, lamm, internvlm, llama_adapter, minigpt4} have demonstrated a general visual ability and can tackle a variety of multi-modality tasks, like visual question answering~\cite{vqav2, mmbench, scienceqa}, document understanding~\cite{chartqa, docvqa, textvqa}, image captioning~\cite{nocaps, cococap, flickr}, \etc
Given MLLMs' advanced performance in these high-level perception tasks, it is natural to explore how MLLMs perform in low-level perception tasks that are highly related to IQA.

\vspace{2pt}\noindent\textbf{MLLM-based IQA methods} leverages MLLMs' foundational knowledge to achieve better IQA performance or more detailed assessment results~\cite{iqasurvey_tianhe, iqasurvey_zicheng}. 
Q-Bench~\cite{qbench, qbench_plus} shows that general MLLMs have some low-level perception ability. 
\cite{2afcprompt} evaluates various MLLMs on the widely-adopted Two-Alternative Forced Choice (2AFC) task. 
Q-Instruct~\cite{qinstruct} enhances the low-level perception ability of MLLMs by introducing a large-scale dataset. 
Co-Instruct~\cite{coinstruct} concentrates on the quality comparison among multiple images. 
DepictQA~\cite{depictqa, depictqav2} performs single-image assessment and paired-image comparison in both full-reference and non-reference settings. 
Q-Ground~\cite{qground} focuses on the visual quality grounding task, and performs fine-scale detailed visual quality analysis. 
Some of these works have proposed MLLM-based quality scoring methods, \eg, binary softmax strategy in~\cite{qbench, qbench_plus} and pair-wise voting in~\cite{depictqa, depictqav2}. 
Nonetheless, these works principally focus on the quality description ability using languages.

\vspace{2pt}\noindent\textbf{MLLM-based quality scorer}. 
Initially, Q-Bench~\cite{qbench, qbench_plus} proposes a binary softmax strategy, enabling MLLMs to generate quality scores by predicting two discrete quality levels (\ie, good or poor). 
This strategy is adopted by Q-Instruct~\cite{qinstruct} and Co-Instruct~\cite{coinstruct}. 
Compare2Score~\cite{compare2score} achieves quality scoring by teaching MLLMs to compare pair-wise images. 
Inspired by humans' annotation process, Q-Align~\cite{qalign} discretizes scores to five discrete levels using one-hot label to train MLLMs, resulting in more accurate score regression. 
Dog-IQA~\cite{dogiqa} adopts the one-hot label for training-free IQA by incorporating specific standards and local semantic objects. 
However, as discussed in~\cref{sec:intro}, one-hot label causes some information loss and can not fully maintain the inter-image relationship.

\section{Soft Label Construction}
\label{sec:soft}

\subsection{Revisiting One-hot Label}
\label{subsec:onehot}

\begin{figure}[t]
    \centering
    \includegraphics[width=0.95\linewidth]{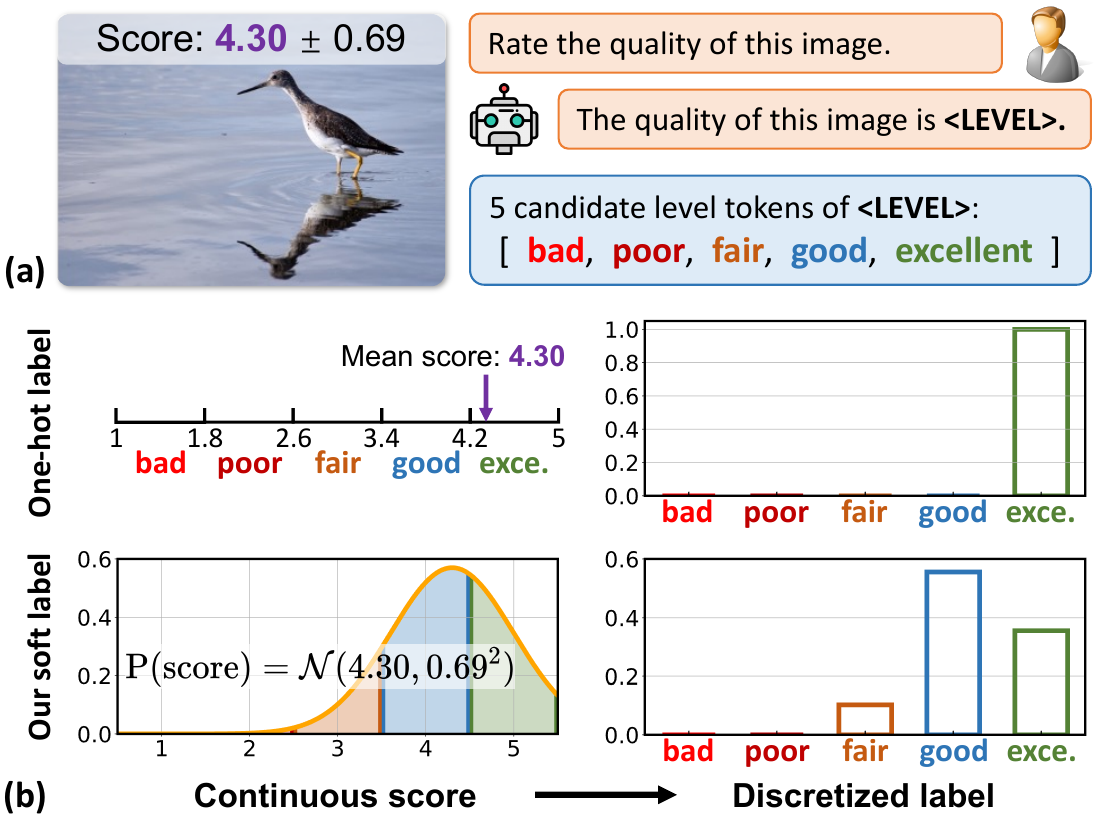}
    \vspace{-5pt}
    \caption{        
    (a) \textbf{MLLMs need discrete level tokens for training}. 
    Thus, continuous scores must be converted into discrete level tokens as training labels.
    (b) \textbf{Comparison between one-hot label~\cite{qalign} and our soft label}. 
    One-hot label uniformly divides the score range into 5 intervals to discretize the mean score, \ie, the expectation of estimated score distribution. 
    One-hot label assumes that different levels are identically independent, which is inaccurate. 
    In contrast, we discretize the estimated Gaussian distribution of the score. 
    Our soft label better preserves the relationships between levels, \eg, ``fair'' is closer to ``good'' than ``excellent''. 
    }
    \label{fig:onehot_soft}
    \vspace{-15pt}
\end{figure}

\textbf{Training with one-hot label}. 
MLLMs only accept discrete tokens as inputs and outputs.
Therefore, as shown in \cref{fig:onehot_soft}a, continuous scores\footnote{Unless otherwise specified, the mean quality scores in all datasets will be normalized to $[1, 5]$, and the variances will be normalized accordingly.} must be discretized into discrete tokens to train the MLLMs. 
Q-Align~\cite{qalign} adopts a one-hot discretization method. 
Specifically, as depicted in \cref{fig:onehot_soft}b, the score range $[1, 5]$ is uniformly divided into five intervals. 
These intervals represent five rating levels, \{``bad'', ``poor'', ``fair'', ``good'', ``excellent''\}, which is the rating standard defined by \cite{five_level}. 
In \cref{fig:onehot_soft}a, the Mean Opinion Score (MOS) of the image, 4.30, fails into the fifth intervals, resulting in the discretized label ``excellent''. 
Accordingly, the response is formulated as ``The quality of this image is excellent''. 
Using the response as the training labels, following~\cite{llama, vicuna}, MLLMs can be trained with next token prediction loss, \ie, cross-entropy loss between predicted logits and labels.

\vspace{2pt}\noindent\textbf{Limitations of one-hot label}. 
First, ideal training with cross-entropy loss using one-hot labels assumes that the five levels are independent (\ie, orthogonal in a latent space). 
However, intuitively, these levels are not truly independent, \eg, ``fair'' is closer to ``good'' than ``excellent''. 
Second, one-hot discretization only considers the mean score, which is the expectation of the score distribution, ignoring the distribution’s uncertainty (\ie, variance). 
Third, near the border of two levels, images with similar quality can be divided into different levels. 
For example, in \cref{fig:teaser}, Image B (MOS 3.38) and Image C (MOS 3.49) share similar mean scores, but are differently classified as ``fair'' and ``good'', due to their proximity to the border of these two levels, \ie, 3.4 in \cref{fig:onehot_soft}b. 
Finally, one-hot label causes information loss. 
In Q-Align, the five levels are respectively assigned scores \{1, 2, 3, 4, 5\}, with which we calculate the discretization precision in \cref{tab:error_discrete}. 
The discretization error is around $10\times$ to $35\times$ larger than our soft label that we will soon introduce, thus the one-hot label keeps lower correlation with MOS.

\begin{table}[t]
\setlength\tabcolsep{1.8pt}
\centering
\footnotesize
\caption{
    \textbf{Comparison of discretization precision} between Q-Align's one-hot label~\cite{qalign} and our soft label. 
    The precision of MOS discretization is measured by L1 Error / RMSE and PLCC / SRCC. 
    The distance of original distribution and recovered distribution is measured by JS divergence / Wasserstein distance. 
}
\vspace{-7pt}
\begin{tabular}{cc|ccc}
\toprule
Metrics & Discretization & KonIQ~\cite{koniq} & SPAQ~\cite{spaq} & KADID~\cite{kadid} \\
\midrule
\multirow{2}{*}{L1 / RSME} &
One-hot label & 0.302 / 0.374 & 0.299 / 0.366 & 0.327 / 0.389 \\
& Soft label & \textbf{0.008} / \textbf{0.018} & \textbf{0.011} / \textbf{0.025} & \textbf{0.024} / \textbf{0.038} \\
\midrule
\multirow{2}{*}{PLCC / SRCC} &
One-hot label & 0.961 / 0.952 & 0.969 / 0.968 & 0.979 / 0.982 \\
& Soft label & \textbf{1.000} / \textbf{1.000} & \textbf{1.000} / \textbf{1.000} & \textbf{1.000} / \textbf{1.000} \\
\midrule
\multirow{2}{*}{JS / W-Dist.} &
One-hot label & - & - & - \\
& Soft label & 0.001 / 0.038 & 0.022 / 0.084 & 0.010 / 0.140 \\
\bottomrule
\end{tabular}
\label{tab:error_discrete}
\vspace{-13pt}
\end{table}

\subsection{Soft Label}

\textbf{Discretization using score distribution}. 
As demonstrated in~\cite{unique}, the quality score of one image, $x$, is typically modeled as a Gaussian distribution, $x \sim \mathcal{N}(\mu, \sigma^2)$, where $\mu$ and $\sigma^2$ are the Mean Opinion Score (MOS) and variance of multiple human annotations, respectively. 
The distribution is represented by its probability density function, $f(x)$. 
As shown in the second row of \cref{fig:onehot_soft}b, we select five central points, $c_i \in \{1,2,3,4,5\}, i \in \{0,1,2,3,4\}$, as the centers of the five discrete levels. 
The width of each level region, $d$, is set as 1. 
The possibility of the score falling into the $i^{\rm th}$ level is denoted as $p_i^{raw}$ and is calculated as: 
\begin{equation}
    p_i^{raw} = \int_{c_i - d/2}^{c_i + d/2}f(x){\rm dx}, \quad i \in \{0,1,2,3,4\}.
    \label{eq:prob_level}
\end{equation}
We assign the five levels with the same textual tokens as Q-Align, \ie, \{``bad'', ``poor'', ``fair'', ``good'', ``excellent''\}. 
Correspondingly, the score of $i^{\rm th}$ level is $c_i$. 
This defines a rough discrete distribution as $P^{raw}(c_i) = p_i^{raw}$.

\vspace{2pt}\noindent\textbf{Post-adjustment}. 
The discrete distribution is rough because of truncation errors. 
As shown in the second row of \cref{fig:onehot_soft}b, there are two truncation regions (\ie, $x < c_0 - d / 2 = 0.5$ and $x > c_4 + d / 2=5.5$), which are not considered in \cref{eq:prob_level}. 
This can cause two problems. 
First, the sum of $\{p_i^{raw}\}$ does not equal 1, meaning $p_i^{raw}$ is not strictly a discrete distribution. 
Second, the expectation of this discrete distribution does not equal the expectation of score distribution, $\mu$, leading to discretization errors.

\begin{figure}
\begin{minipage}[t]{1.0\columnwidth}
\begin{minipage}[t]{0.405\textwidth}
    \centering
    \includegraphics[width=1.0\columnwidth]{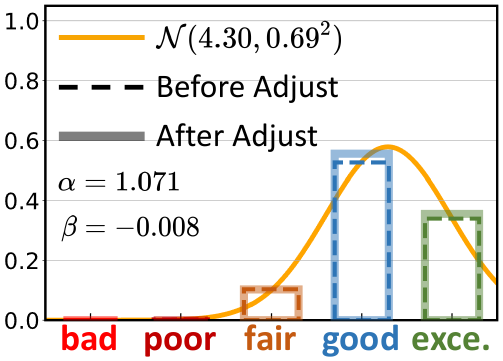}
    \vspace{-18pt}
    \captionof{figure}{
    \textbf{Illustration of the soft label adjustment}. 
    }
    \label{fig:adjust}
\end{minipage}
\hfill
\begin{minipage}[t]{0.58\columnwidth}
    \setlength\tabcolsep{1.5pt}
    \centering
    \scriptsize
    \vspace{-70pt}
    \captionof{table}{
        \textbf{Statistics of the label adjustment parameters}. 
        Averaged $\alpha$ and $\beta$ are very approaching 1 and 0, respectively, meaning that the post-adjustment is quite small. 
    }
    \vspace{-7pt}
    \begin{tabular}{c|ccc}
    \toprule
    & KonIQ~\cite{koniq} & SPAQ~\cite{spaq} & KADID~\cite{kadid} \\
    \midrule
    $\alpha$ & 1.036$\pm$0.041 & 1.033$\pm$0.085 & 1.118$\pm$0.091 \\
    $\beta$ & -0.005$\pm$0.005 & -0.005$\pm$0.014 & -0.015$\pm$0.010 \\
    \bottomrule
    \end{tabular}
    \label{tab:adjust}
\end{minipage}
\end{minipage}
\vspace{-14pt}
\end{figure}

To correct these truncation errors, we propose a post-adjustment method. 
Specifically, we apply a linear transformation to adjust $p_i^{raw}$ to more accurate $p_i$ as: 
\begin{align}
    p_i &= \alpha p_i^{raw} + \beta \nonumber \\
    {\rm s.t.} \quad \sum\nolimits_i p_i &= 1, \quad \mu^{rec} = \sum\nolimits_i p_i c_i = \mu, \label{eq:adjust}
\end{align}
where $\mu_{rec}$ means recovered expectation from discrete distribution. 
The parameters of the linear transformation, $\alpha$ and $\beta$, can be determined by solving these two constraint conditions.
One example of post-adjustment is given in \cref{fig:adjust}. 
Moreover, the statistics of these two parameters in \cref{tab:adjust} show that the average values of $\alpha$ and $\beta$ are quite close to 1 and 0, respectively, implying that the post-adjustment is minimal. 
This effectively defines our soft label, which follows the discrete distribution as $P(c_i) = p_i$.

Ideally, the discretization error of the MOS will be 0. 
However, the possibility of some levels may slightly fall below zero after the linear transformation. 
We address this by simply clipping these values to zero.
The impact of this clipping on accuracy is negligible, as shown in \cref{tab:error_discrete}.
After clipping, our discretization error for MOS remains around $10\times$ to $35\times$ smaller than Q-Align's one-hot label. 
Also, our soft label maintains a perfect (\ie, 1.000) Pearson Linear Correlation Coefficient (PLCC) and Spearman Rank-order Correlation Coefficient (SRCC) with the original MOS.

\vspace{2pt}\noindent\textbf{Recovery from discrete to continuous score distribution}. 
As shown in the second condition of \cref{eq:adjust}, we can compute the expectation of discrete distribution, $P(c_i)$, as the recovered expectation of original distribution. 
Besides the expectation, we can also recover the original variance using the variance of the discrete distribution, $P(c_i)$, as: 
\begin{equation}
    \mu^{rec} = \sum\nolimits_i p_i c_i,
    \quad 
    (\sigma^{rec})^2 = \sum\nolimits_i p_i (c_i - \mu^{rec})^2.
    \label{eq:rec}
\end{equation}
Then we calculate the distance between recovered distribution, $\mathcal{N}(\mu^{rec}, (\sigma^{rec})^2)$, and original distribution, $\mathcal{N}(\mu, \sigma^2)$, as provided in \cref{tab:error_discrete}. 
Our recovered distribution is closely aligned with human-annotated score distribution. 
In contrast, one-hot label can not recover such a distribution.

\begin{figure*}[t]
    \centering
    \includegraphics[width=0.95\linewidth]{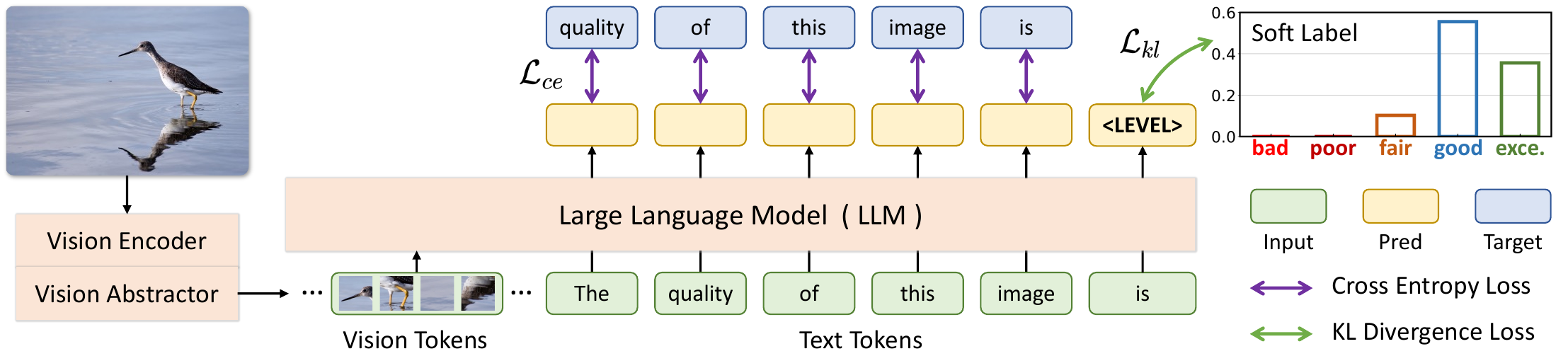}
    \vspace{-7pt}
    \caption{
    \textbf{Framework of our \method trained with soft label}. 
    For the ``\(<\)level\(>\)'' token, the KL divergence loss is calculated between predicted probabilities and our soft label. For other tokens, common cross-entropy loss for next token prediction is calculated. 
    }
    \label{fig:model}
    \vspace{-15pt}
\end{figure*}

\section{\method Model}

\subsection{Training with Our Soft Label}

\textbf{Model architecture}. 
As illustrated in \cref{fig:model}, we follow the widely adopted MLLM architecture~\cite{llava, mplugowl2} to construct our model. 
First, a vision encoder is adopted to represent the input image to visual tokens. 
Then, a vision-to-text connector projects the visual tokens into the textual space. 
Following~\cite{mplugowl2}, we use a vision abstractor as the connector. 
The abstractor can further reduce the number of visual tokens to 64. 
Finally, the visual and textual tokens are fused together and fed into an LLM for response prediction.

\vspace{2pt}\noindent\textbf{Next token prediction loss for common response tokens}. 
As described in \cref{subsec:onehot} and illustrated in \cref{fig:model}, the constructed response follows the template, “The quality of this image is \(<\)level\(>\)”. 
For common response tokens other than the level token, we compute the next token prediction loss, \ie, the cross-entropy loss \( \mathcal{L}_{ce} \), following~\cite{llama, gpt3.5}.

\vspace{2pt}\noindent\textbf{KL divergence loss for the level token}. 
There are five candidate values for the level token: \{``bad'', ``poor'', ``fair'', ``good'', ``excellent''\}. 
Our soft label, $p_i$, assigns a ground-truth probability to each value.
Let the predicted probabilities be $p_i^{pred}$, the KL divergence loss is calculated through: 
\begin{equation}
    \mathcal{L}_{kl} = - \sum\nolimits_i p_i \log (p_i^{pred} / p_i).
\end{equation}
The predicted probabilities, $p_i^{pred}$, are obtained by applying a softmax activation function on model's output logits. 
Note that during training, this softmax function is applied across all textual tokens, not just across the five predefined levels. 
This approach encourages the model to reduce the possibilities of textual tokens outside the five levels.

The two terms of losses, $\mathcal{L}_{ce}$ and $\mathcal{L}_{kl}$, are both auto-regressive losses, and thus can be easily combined. 
During training, the cross-entropy loss, $\mathcal{L}_{ce}$, will decrease near to 0 fast, since it is quite easy to optimize by simply remembering the response template ``The quality of this image is''.

\subsection{Score Estimation during Inference}
\label{subsec:infer}

During inference, using the model-predicted probabilities of the five levels, $p_i^{pred}$, we can estimate the expectation and variance of the score distribution the same as \cref{eq:rec}:
{{
\footnotesize
\begin{equation}
    \mu^{pred} = \sum\nolimits_i p_i^{pred} c_i, \quad
    (\sigma^{pred})^2 = \sum\nolimits_i p_i^{pred} (c_i - \mu^{pred})^2.
    \label{eq:pred}
\end{equation}
}}
Here we follow Q-Align~\cite{qalign} to apply a close-set softmax on the five levels to obtain the probabilities, $p_i^{pred}$, to prevent the influence of other textual tokens. 
Our statistics in \cref{supp:tab:prob_sum} reveal that the trained MLLM will always predict the five levels, thus applying softmax on either the five levels or all textual tokens yields nearly identical results. 
As given in \cref{tab:pred_dist}, our predicted score distribution, $\mathcal{N}(\mu^{pred}, (\sigma^{pred})^2)$, aligns well with human annotations.

\begin{figure}[t]
    \centering
    \includegraphics[width=0.95\linewidth]{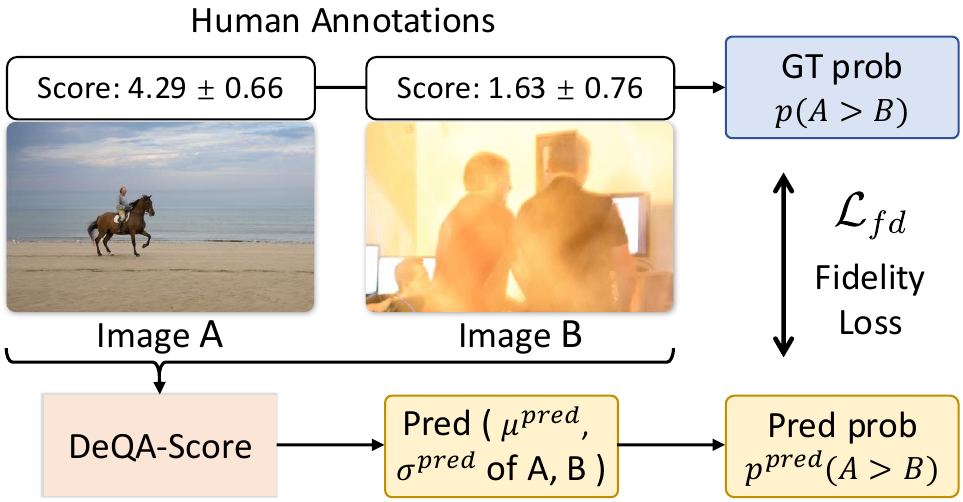}
    \vspace{-5pt}
    \caption{
    \textbf{Fidelity loss to maintain intra-dataset relationships}. 
    Consider two images, Image A and B, sampled from the same dataset. 
    Using our distribution-based soft label, \method predicts the expectation, $\mu^{pred}$, and variance, $(\sigma^{pred})^2$, of each image’s score distribution, with which we can predict the probability that Image A is better than Image B, $P(A>B)$. 
    This predicted probability is then used to calculate the fidelity loss with the ground-truth probability. 
    Fidelity loss preserves intra-dataset relationships, facilitating the co-training across multiple IQA datasets. 
    }
    \label{fig:fidelity}
    \vspace{-15pt}
\end{figure}

\subsection{Joint Training on Multiple IQA Datasets}

As demonstrated in~\cite{unique, compare2score}, different IQA datasets exhibit distinct distributions. 
Due to this dataset variation, images from different datasets with nearly identical scores (\ie, linearly re-scaled scores) can have drastically different perceptual quality (see examples in \cref{supp:fig:dataset}). 
The reason is that different datasets have different reflections between quality scores and visual quality. 
This disparity makes it infeasible to merely regress absolute quality scores for multi-dataset co-training. 
To address this challenge, UNIQUE~\cite{unique} applies fidelity loss~\cite{fidelity} based on Thurstone's model~\cite{thurstone}. 
The fidelity loss is based on the principal that IQA emphasizes the ranking relationships within the same dataset, rather than the absolute values of quality scores. 
Specifically, two images are sampled from the same dataset, whose quality scores can be comparable. 
These two images are then compared using human-annotated quality scores. 
The fidelity loss encourages the model to learn such comparison results, which facilitates the model to focus more on the ranking relationships in each dataset, instead of the absolute values of quality scores. 
Thus, the fidelity loss helps multi-dataset co-training. 
Our soft label is naturally compatible with the fidelity loss, since our method can predict the score distribution, which is what the fidelity loss needs.

\vspace{2pt}\noindent\textbf{Ground-truth construction}. 
As shown in \cref{fig:fidelity}, a pair of images, Image A and B, are sampled from the same dataset. 
We assume that the quality scores of the two images, $x_A, x_B$, follow the Gaussian distribution with means and variances collected by subjective testing, \ie, $x_A \sim \mathcal{N}(\mu_A, \sigma_A^2), x_B \sim \mathcal{N}(\mu_B, \sigma_B^2)$. 
Assuming that their quality scores are uncorrelated, the score difference, $x_A-x_B$, also follows a Gaussian distribution, $\mathcal{N}(\mu_A-\mu_B, \sigma_A^2+\sigma_B^2)$. 
Then the possibility that Image A has a better perceptual quality than Image B can be calculated as: 
\begin{equation}
    p(A>B) = p(x_A>x_B) = \Phi \left( \frac{\mu_A-\mu_B}{\sqrt{\sigma_A^2+\sigma_B^2}} \right), 
    \label{eq:gt_AB}
\end{equation}
where $\Phi(\cdot)$ is the Gaussian cumulative distribution function.

\vspace{2pt}\noindent\textbf{Model prediction}. 
As described in \cref{subsec:infer}, our model can predict the expectation and variance of the score distribution. 
Denoting the predicted expectation and variance of two images as $\mu_A^{pred}$, $(\sigma_A^{pred})^2$, $\mu_B^{pred}$, $(\sigma_B^{pred})^2$, same to \cref{eq:gt_AB}, we can predict the possibility of Image A surpassing Image B in perceptual quality as: 
\begin{equation}
    p^{pred}(A>B) = \Phi \left( \frac{\mu_A^{pred} - \mu_B^{pred}}{\sqrt{(\sigma_A^{pred})^2 + (\sigma_B^{pred})^2}} \right).
\end{equation}

\vspace{2pt}\noindent\textbf{Training with fidelity loss}. 
Following~\cite{unique}, we choose fidelity loss as the similarity measure to train the model as: 
\begin{align}    
    \mathcal{L}_{fd} & = 1 - \sqrt{p(A>B)p^{pred}(A>B)} \nonumber \\
    &- \sqrt{\left( 1-p(A>B) \right) \left( 1-p^{pred}(A>B) \right) }. 
\end{align}
Adding two auto-regressive losses, the final training loss is: 
\begin{equation}
    \mathcal{L} = \mathcal{L}_{fd} + \gamma (\mathcal{L}_{ce} + \mathcal{L}_{kl}), 
\end{equation}
where $\gamma$ is a weighting term to adjust the loss scale.


\begin{table*}[t]
\setlength\tabcolsep{2.6pt}
\centering
\footnotesize
\caption{
    \textbf{Score regression results of single-dataset training} with the PLCC / SRCC metrics. 
    All methods except handcrafted ones are trained on the KonIQ dataset. 
    \method outperforms all baseline methods across nearly all benchmarks. 
}
\vspace{-7pt}
\label{tab:single_data}
\begin{tabular}{c|c|c|ccccccc}
\toprule
Category & Methods & KonIQ & SPAQ & KADID & PIPAL & LIVE-Wild & AGIQA-3K & CSIQ & FLIVE \\
\midrule
\multirow{2}{*}{Handcrafted} 
& NIQE~\cite{niqe} & 
0.533 / 0.530 & 
0.679 / 0.664 &
0.468 / 0.405 &
0.195 / 0.161 &
0.493 / 0.449 &
0.560 / 0.533 &
0.718 / 0.628 & 
0.147 / 0.100 \\
& BRISQUE~\cite{brisque} & 
0.225 / 0.226 & 
0.490 / 0.406 & 
0.429 / 0.356 & 
0.267 / 0.232 &
0.361 / 0.313 &
0.541 / 0.497 &
0.740 / 0.556 &
0.108 / 0.054 \\
\midrule
\multirow{6}{*}{\makecell[c]{Non-MLLM \\ Deep-learning}} 
& NIMA~\cite{nima} & 
0.896 / 0.859 & 
0.838 / 0.856 & 
0.532 / 0.535 & 
0.390 / 0.399 & 
0.814 / 0.771 & 
0.715 / 0.654 & 
0.695 / 0.649 &
0.561 / 0.467 \\
& HyperIQA~\cite{HyperIQA} & 
0.917 / 0.906 &
0.791 / 0.788 &
0.506 / 0.468 &
0.410 / 0.403 & 
0.772 / 0.749 &
0.702 / 0.640 & 
0.752 / 0.717 &
0.485 / 0.383 \\
& DBCNN~\cite{dbcnn} & 
0.884 / 0.875 &
0.812 / 0.806 &
0.497 / 0.484 & 
0.384 / 0.381 &
0.773 / 0.755 &
0.730 / 0.641 &
0.586 / 0.572 &
0.485 / 0.385 \\
& MUSIQ~\cite{musiq} & 
0.924 / 0.929 & 
0.868 / 0.863 &
0.575 / 0.556 & 
0.431 / 0.431 &
0.789 / 0.830 &
0.722 / 0.630 &
0.771 / 0.710 &
0.565 / 0.467 \\
& CLIP-IQA+~\cite{clipiqa} & 
0.909 / 0.895 &
0.866 / 0.864 &
0.653 / 0.654 & 
0.427 / 0.419 &
0.832 / 0.805 &
0.736 / 0.685 &
0.772 / 0.719 &
0.427 / 0.316 \\
& ManIQA~\cite{maniqa} & 
0.849 / 0.834 &
0.768 / 0.758 &
0.499 / 0.465 & 
0.457 / 0.452 &
0.849 / 0.832 &
0.723 / 0.636 &
0.623 / 0.627 &
0.512 / 0.401 \\
\midrule
\multirow{3}{*}{MLLM-based} 
& {\scriptsize Compare2Score}~\cite{compare2score} & 
0.923 / 0.910 &
0.867 / 0.860 &
0.500 / 0.453 &
0.354 / 0.342 &
0.786 / 0.772 &
0.777 / 0.671 &
0.735 / 0.705 & 
0.474 / 0.413 \\
& Q-Align~\cite{qalign} & 
0.941 / 0.940 &
0.886 / 0.887 &
0.674 / 0.684 & 
0.403 / 0.419 &
0.853 / 0.860 &
0.772 / \textbf{0.735} &
0.785 / 0.737 &
0.554 / 0.483 \\
& \method & 
\textbf{0.953} / \textbf{0.941} & 
\textbf{0.895} / \textbf{0.896} & 
\textbf{0.694} / \textbf{0.687} & 
\textbf{0.472} / \textbf{0.478} & 
\textbf{0.892} / \textbf{0.879} & 
\textbf{0.809} / 0.729 & 
\textbf{0.787} / \textbf{0.744} &
\textbf{0.589} / \textbf{0.501} \\
\bottomrule
\end{tabular}
\vspace{-7pt}
\end{table*}

\begin{table*}[t]
\setlength\tabcolsep{3pt}
\centering
\footnotesize
\caption{
    \textbf{Score regression results of co-training on multiple IQA datasets} with the PLCC / SRCC metrics. 
    Our \method consistently surpasses the baseline method, Q-Align, across all benchmarks with different training combinations. 
}
\vspace{-7pt}
\label{tab:multi_data}
\begin{tabular}{c|c|c|cccccccc}
\toprule
\# & Trained on & Methods & KonIQ & SPAQ & KADID & PIPAL & LIVE-Wild & AGIQA-3K & TID2013 & CSIQ \\
\midrule
\multirow{2}{*}{0} & \multirow{2}{*}{{\scriptsize KonIQ, SPAQ}}
& Q-Align~\cite{qalign} & 
0.943 / 0.940 & 
0.933 / 0.931 & 
0.692 / 0.708 & 
0.401 / 0.411 & 
0.883 / 0.879 & 
0.795 / 0.727 & 
0.671 / 0.568 & 
0.795 / 0.767 \\
& & \method & 
\textbf{0.953} / \textbf{0.943} & 
\textbf{0.936} / \textbf{0.933} & 
\textbf{0.724} / \textbf{0.719} & 
\textbf{0.468} / \textbf{0.474} & 
\textbf{0.902} / \textbf{0.888} & 
\textbf{0.810} / \textbf{0.738} & 
\textbf{0.706} / \textbf{0.616} & 
\textbf{0.836} / \textbf{0.785} \\
\midrule
\multirow{2}{*}{1} & \multirow{2}{*}{\makecell[c]{\scriptsize KonIQ, SPAQ,\\ \scriptsize KADID}}
& Q-Align~\cite{qalign} & 
0.945 / 0.938 & 
0.933 / 0.931 & 
0.935 / 0.934 & 
0.409 / 0.420 & 
0.887 / 0.883 & 
0.788 / 0.733 & 
0.829 / 0.808 & 
0.876 / 0.845 \\
& & \method & 
\textbf{0.957} / \textbf{0.944} & 
\textbf{0.938} / \textbf{0.934} & 
\textbf{0.955} / \textbf{0.953} & 
\textbf{0.495} / \textbf{0.496} & 
\textbf{0.900} / \textbf{0.887} & 
\textbf{0.808} / \textbf{0.745} & 
\textbf{0.852} / \textbf{0.820} & 
\textbf{0.900} / \textbf{0.857} \\
\midrule
\multirow{2}{*}{2} & \multirow{2}{*}{\makecell[c]{\scriptsize KonIQ, SPAQ,\\ \scriptsize KADID, PIPAL}}
& Q-Align~\cite{qalign} & 
0.926 / 0.932 &
0.917 / 0.920 &
0.950 / 0.954 &
0.702 / 0.671 &
0.853 / 0.845 &
0.765 / 0.722 &
0.811 / 0.795 &
0.838 / 0.789 \\
& & \method & 
\textbf{0.958} / \textbf{0.946} &
\textbf{0.932} / \textbf{0.929} &
\textbf{0.963} / \textbf{0.961} &
\textbf{0.724} / \textbf{0.690} &
\textbf{0.877} / \textbf{0.857} &
\textbf{0.770} / \textbf{0.735} &
\textbf{0.828} / \textbf{0.796} &
\textbf{0.863} / \textbf{0.807} \\
\bottomrule
\end{tabular}
\vspace{-14pt}
\end{table*}

\section{Experiments}
\label{sec:exp}

\subsection{Datasets, Baselines, Metrics, and Details}

\textbf{Datasets}. 
For a comprehensive evaluation, we include a wide range of IQA datasets: 
(a) in-the-wild images: KonIQ~\cite{koniq}, SPAQ~\cite{spaq}, LIVE-Wild~\cite{livewild}, and FLIVE~\cite{paq2piq};
(b) synthetic distortions: KADID~\cite{kadid}, TID2013~\cite{tid2013}, and CSIQ~\cite{csiq}; 
(c) model-processed distortions: PIPAL~\cite{pipal}; 
(d) AI-generated images: AGIQA-3K~\cite{agiqa}. 
Among these datasets, KonIQ, SPAQ, KADID, and PIPAL are split into training and test sets. 
Other datasets are used to assess the out-of-distribution generalization ability. 
The Mean Opinion Scores (MOS) of these datasets are normalized to the range of [1,5], with the variances normalized accordingly.

\vspace{2pt}\noindent\textbf{Baseline methods}. 
To demonstrate the effectiveness of our distribution-based soft label, we primarily compare with Q-Align~\cite{qalign}, which utilizes the one-hot label. 
The base models for \method and Q-Align are kept identical for fair comparison. 
We also include previous IQA methods as baselines, including handcrafted methods: NIQE~\cite{niqe} and BRISQUE~\cite{brisque}, and deep-learning-based methods: NIMA~\cite{nima}, HyperIQA~\cite{HyperIQA}, DBCNN~\cite{dbcnn}, MUSIQ~\cite{musiq}, CLIP-IQA+~\cite{clipiqa}, ManIQA~\cite{maniqa}, and Compare2Score~\cite{compare2score}.

\vspace{2pt}\noindent\textbf{Metrics}. 
Following~\cite{musiq, qalign, maniqa}, we use the Pearson Linear Correlation Coefficient (PLCC) and Spearman Rank-order Correlation Coefficient (SRCC) as metrics to evaluate the score regression performance. 
Additionally, we employ KL divergence, JS divergence, and Wasserstein distance to assess the results of score distribution prediction.

\vspace{2pt}\noindent\textbf{Implementation details}. 
We use mPLUG-Owl2~\cite{mplugowl2} as our base model, with a CLIP-pretrained ViT-L~\cite{clip} as the image encoder, a six-layer Q-Former as the visual abstractor, and LLaMA-2-7B~\cite{llama2} as the LLM. 
The number of visual tokens output by the vision abstractor is set to 64, and the hidden dimension of LLM is 4096. 
The pretrained weights of mPLUG-Owl2 are used for model initialization. 
The weighting term $\gamma$ is empirically set to 0.05, adjusting the two loss terms roughly to the same scale. 
We adopt AdamW~\cite{adamw} as the optimizer. 
The initial learning rate is set to 2e-5 and decays gradually using the cosine decay strategy.
Our model is trained with a batch size of 64 for 3 epochs.
Using 8 NVIDIA RTX A6000 GPUs, the training process on KonIQ dataset is completed in around 1.5 hours.

\subsection{Results of Score Regression}
\label{subsec:exp_single}

\textbf{Score regression results of single-dataset training}. 
We first train our model on a single dataset, \ie, KonIQ, and then evaluate the model on both in-distribution and out-of-distribution datasets, following~\cite{attiqa, qalign}. 
The score regression results are shown in \cref{tab:single_data}.
On the in-distribution KonIQ test set, our model achieves superior score regression results compared to all baseline methods, \eg, a 1.3\% improvement in PLCC over Q-Align. 
Furthermore, on out-of-distribution datasets, our \method maintains its advantage with the best overall results across nearly all benchmarks, \eg, a 4.6\% improvement in PLCC on LIVE-Wild. 
These findings confirm the effectiveness of our method.

However, we observe that all models perform poorly on the PIPAL dataset.
For instance, our best results reach only 0.462 on PLCC.
The main reason is the distortion gap between the training dataset, KonIQ (in-the-wild distortions), and PIPAL (model-processed distortions). 
This highlights the necessity of co-training with multiple IQA datasets containing various distortion types.
This observation motivates our subsequent experiments on multi-dataset co-training.

\vspace{2pt}\noindent\textbf{Score regression results of multi-dataset co-training}. 
As discussed above, co-training on multiple IQA datasets is necessary to obtain a unified model for various categories of distortions. 
Therefore, we conduct multi-dataset co-training experiments in \cref{tab:multi_data}. 
First, our \method consistently outperforms Q-Align across different training combinations. 
The performance gain attributes to both the soft label and fidelity loss, as confirmed in \cref{tab:ablation_design}. 
Second, comparing \#0 to \#1, when KADID (synthetic distortions) is added to the training set, the performance on TID2013 and CSIQ (both synthetic distortions) improves significantly (\eg, 33.1\% in SRCC on TID2013), underscoring that distortion type is a major gap between training and testing. 
Third, from \#1 to \#2, adding PIPAL for training greatly enhances the results on PIPAL (from 0.495 to 0.724 in PLCC) but reduces the performance on several other datasets, \eg, from 0.900 to 0.863 in PLCC on CSIQ. 
This performance decline is likely due to the unique nature of model-processed distortions in PIPAL, which differ greatly from other distortions. 
Finally, despite never trained on AI-generated images, \method performs well on AGIQA-3K with around 0.8 PLCC, reflecting the generalization ability enabled by MLLMs' broad pre-training.

\vspace{2pt}\noindent\textbf{More results on AIGC images} are provided in \cref{supp:tab:res_aigc}.

\subsection{Results of Score Distribution Prediction}
\label{subsec:dist_pred}

\begin{table}[t]
\setlength\tabcolsep{12pt}
\centering
\small
\caption{
    \textbf{Score distribution prediction results} with JS divergence / Wasserstein distance between the predicted Gaussian distribution and the human labeled Gaussian distribution as metrics. 
    Models are trained on the KonIQ, SPAQ, and KADID datasets. 
    \method achieves a much closer alignment with human annotations. 
}
\vspace{-8pt}
\begin{tabular}{c|cc}
\toprule
Datasets & Q-Align~\cite{qalign} & \method (Ours) \\
\midrule
KonIQ & 0.415 / 0.611 & \textbf{0.014} / \textbf{0.186} \\
SPAQ  & 0.192 / 0.455 & \textbf{0.070} / \textbf{0.332} \\
KADID & 0.519 / 0.723 & \textbf{0.029} / \textbf{0.304} \\
LIVE-Wild & 0.254 / 0.642 & \textbf{0.052} / \textbf{0.417} \\
AGIQA-3K & 0.558 / 0.730 & \textbf{0.177} / \textbf{0.592} \\
\bottomrule
\end{tabular}
\label{tab:pred_dist}
\vspace{-15pt}
\end{table}

\begin{figure*}[t]
    \centering
    \includegraphics[width=0.95\linewidth]{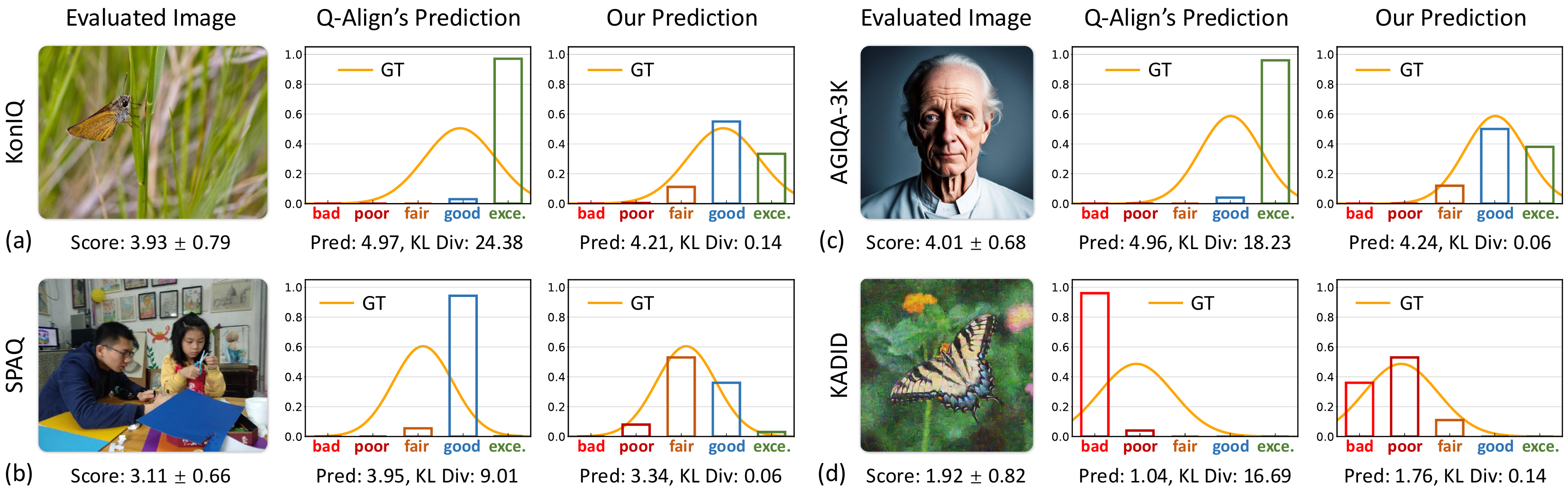}
    \vspace{-7pt}
    \caption{
    \textbf{Qualitative results} on KonIQ, SPAQ, AGIQA-3K, and KADID datasets. 
    \method predicts more accurate quality scores with more reasonable logits.
    Also, our method generates a score distribution that aligns better with humans, \ie, smaller KL divergence. 
    However, Q-Align, which is trained on one-hot label, is more likely to predict a single label, resulting in higher KL divergence.  
    }
    \label{fig:res}
    \vspace{-7pt}
\end{figure*}

\begin{table*}[t]
\setlength\tabcolsep{1.8pt}
\centering
\footnotesize
\captionof{table}{
    \textbf{Ablation studies} on soft label and fidelity loss with PLCC / SRCC metrics. Models are trained on KonIQ, SPAQ, and KADID. 
}
\vspace{-7pt}
\label{tab:ablation_design}
\begin{tabular}{c|cc|cc|ccc|ccccc}
\toprule
\# & One-hot & Soft & Uncertainty & Fidelity & KonIQ & SPAQ & KADID & PIPAL & LIVE-Wild & AGIQA-3K & TID2013 & CSIQ \\
\midrule
0 & \ding{52} & & & & 
0.945 / 0.938 & 
0.933 / 0.931 & 
0.935 / 0.934 & 
0.409 / 0.420 & 
0.887 / 0.883 & 
0.788 / 0.733 & 
0.829 / 0.808 & 
0.876 / 0.845 \\
1 & \ding{52} & & \ding{52} & & 
0.935 / 0.936 &
0.933 / 0.931 &
0.954 / 0.951 &
0.439 / 0.453 &
0.896 / \textbf{0.887} &
0.764 / 0.715 &
0.826 / 0.803 &
0.884 / 0.849 \\
2 & \ding{52} & & & \ding{52} & 
0.944 / 0.936 & 
0.935 / 0.932 & 
0.951 / 0.950 &
0.473 / 0.478 &
0.898 / 0.886 &
0.788 / 0.726 &
0.837 / 0.810 &
0.876 / 0.839 \\
3 & & \ding{52} & & & 
0.954 / 0.943 & 
0.937 / \textbf{0.934} &
0.951 / 0.947 &
0.472 / 0.472 &
0.896 / 0.879 &
0.791 / 0.729 &
\textbf{0.852} / \textbf{0.821} &
0.894 / 0.855 \\
4 & & \ding{52} & & \ding{52} & 
\textbf{0.957} / \textbf{0.944} & 
\textbf{0.938} / \textbf{0.934} & 
\textbf{0.955} / \textbf{0.953} & 
\textbf{0.495} / \textbf{0.496} & 
\textbf{0.900} / \textbf{0.887} & 
\textbf{0.808} / \textbf{0.745} & 
\textbf{0.852} / 0.820 & 
\textbf{0.900} / \textbf{0.857} \\
\bottomrule
\end{tabular}
\vspace{-3pt}
%
\setlength\tabcolsep{2pt}
\centering
\footnotesize
\captionof{table}{
    \textbf{Effects of different level texts} with PLCC / SRCC metrics. Models are trained on the KonIQ, SPAQ, and KADID datasets. 
}
\vspace{2pt}
\label{tab:level_text}
\begin{tabular}{c|c|ccc|ccccc}
\toprule
Category & Level Texts & KonIQ & SPAQ & KADID & PIPAL & LIVE-Wild & AGIQA-3K & TID2013 & CSIQ \\
\midrule
Common & {\scriptsize exce. / good / far / poor / bad} & 
\textbf{0.957} / \textbf{0.944} & 
\textbf{0.938} / \textbf{0.934} & 
\textbf{0.955} / \textbf{0.953} & 
\textbf{0.495} / \textbf{0.496} & 
\textbf{0.900} / \textbf{0.887} & 
\textbf{0.808} / \textbf{0.745} & 
0.852 / 0.820 & 
\textbf{0.900} / \textbf{0.857} \\
Random & {\scriptsize exce. / poor / bad / good / fair} & 
0.942 / 0.928 & 
0.933 / 0.929 &
0.950 / 0.947 &
0.394 / 0.396 &
0.863 / 0.851 &
0.795 / 0.727 &
0.814 / 0.792 &
0.887 / 0.836 \\
Reverse & {\scriptsize bad / poor / fair / good / exce.} & 
0.943 / 0.934 &
0.934 / 0.931 &
0.947 / 0.945 &
0.395 / 0.386 &
0.836 / 0.800 &
0.782 / 0.714 &
0.811 / 0.791 &
0.882 / 0.834 \\
Number & {\scriptsize one / two / three / four / five} & 
0.945 / 0.932 &
0.934 / 0.930 &
0.955 / 0.952 &
0.421 / 0.424 &
0.873 / 0.857 &
0.766 / 0.712 &
0.820 / 0.794 &
0.882 / 0.835 \\
Vehicle & {\scriptsize car / bus / train / plane / ship} & 
0.953 / 0.940 &
0.937 / 0.933 &
0.954 / 0.951 &
0.458 / 0.456 &
0.889 / 0.878 &
0.778 / 0.702 &
\textbf{0.854} / \textbf{0.828} &
0.897 / 0.856 \\
Mixture & {\scriptsize apple / car / four / fair / dog} & 
0.947 / 0.938 &
0.936 / 0.932 &
0.954 / 0.951 &
0.419 / 0.415 &
0.865 / 0.851 &
\textbf{0.808} / 0.737 &
0.827 / 0.800 &
0.896 / 0.854 \\
\bottomrule
\end{tabular}
\vspace{-10pt}
\end{table*}

As stated in \cref{subsec:infer}, the MLLM-output probabilities allow us to predict a score distribution, which can be evaluated against the human labeled ground-truth distribution. 
We calculate the distance between these two distributions and present the results in \cref{tab:pred_dist}. 
The results illustrate that our predicted score distribution aligns much better with humans than Q-Align. 
As shown in \cref{fig:res}, Q-Align tends to predict a single level token, bringing a quite small predicted variance, which compromises the distribution metrics.

\vspace{2pt}\noindent\textbf{Qualitative results} are shown in \cref{fig:res}. 
Across different image types: \textit{in-the-wild} images (\cref{fig:res}ab), \textit{AI-generated} images (\cref{fig:res}c), or \textit{synthetic} distortions (\cref{fig:res}d), 
and for varying quality levels: \textit{high-quality} (\cref{fig:res}ac), \textit{fair-quality} (\cref{fig:res}b), or \textit{poor-quality} images (\cref{fig:res}d), our method consistently predicts accurate quality scores. 
Additionally, our \method provides a score distribution that aligns more closely with humans with smaller KL divergence. 
In contrast, Q-Align, which is trained with one-hot labels, tends to predict a single label, resulting in higher KL divergence.

\subsection{Ablation Studies}

\begin{table}[t]
\setlength\tabcolsep{2.5pt}
\centering
\footnotesize
\vspace{-2pt}
\caption{
    \textbf{Inference latency \& throughput} on RTX A6000 GPU. 
}
\vspace{-8pt}
\begin{tabular}{c|ccccccc}
\toprule
Batch Size & 1 & 2 & 4 & 8 & 16 & 64 & 256 \\
\midrule
Latency (\textit{s}) & 0.089 & 0.121 & 0.182 & 0.315 & 0.588 & 2.114 & 7.929 \\
Throughput (\textit{image/s}) & 11.24 & 16.53 & 21.98 & 25.40 & 27.21 & 30.27 & 32.29 \\
\bottomrule
\end{tabular}
\label{tab:infer_time}
\vspace{-15pt}
\end{table}

\textbf{Effectiveness of soft label and fidelity loss} is studied and shown in \cref{tab:ablation_design}. 
In \#1, ``Uncertainty'' is utilized to verify the effects of adding annotation variances without considering the score distortion. 
We directly append uncertainty within the responses, \ie, ``The quality of this image is \(<\)level\(>\) with \(<\)uncertainty\_level\(>\) uncertainty''. 
The \(<\)uncertainty\_level\(>\) has five possible values: \{``minimal'', ``low'', ``moderate'', ``high'', ``severe''\}, derived by the one-hot discretization. 
In \#2, model trained on one-hot label can not predict a reliable variance, thus we employ binary fidelity loss~\cite{liqe}, which does not require predicted variance.

Several conclusions can be drawn from \cref{tab:ablation_design}. 
First, comparing \#0 with \#3, our soft label significantly outperforms the one-hot label across all benchmarks, validating its effectiveness. 
Second, from \#3 to \#4, the performance of our soft label further improves when combined with the fidelity loss. 
Third, comparing \#0 with \#2, the one-hot label does not incorporate well with the fidelity loss, as it lacks score distribution information. 
Finally, from \#0 to \#1, simply adding uncertainty without considering score distribution does not yield stable or significant improvements.

\vspace{2pt}\noindent\textbf{Effects of different level texts} are studied in \cref{tab:level_text}. 
Beside the commonly used level texts, we consider five alternatives: 
(a) random sorted common texts; 
(b) reversed common texts; 
(c) simple numbers like one, two, \etc; 
(d) irrelevant words about vehicle; 
(e) a mixture of various types. 
The results indicate that common level texts achieve the best performance, as they align closely with the MLLMs’ pre-training. 
Interestingly, with random sorted / reversed texts, even a mixture of various level texts, which are inconsistent with MLLMs' prior knowledge, our model still learns these newly defined relationships and performs well. 
This shows the robust adaptability of MLLM-based IQA models.

\vspace{2pt}\noindent\textbf{Ablation studies on level number and training / fixing components} are provided in \cref{supp:tab:ablation_level_number,supp:tab:ablation_train}, respectively.

\subsection{Inference Latency}

We test the inference latency on one single RTX A6000 GPU. 
As shown in \cref{tab:infer_time}, with the optimal GPU utilization (\ie, batch size 256), our \method can process as many as 32.29 images per second in average. 
This high throughput emphasizes its suitability for real-time deployment.

\section{Conclusions}
\label{sec:conclusions}

We develop \method as a pioneering MLLM-based image quality scorer, empowered by distribution-based soft labels and fidelity loss, capable of both score regression and score distribution prediction, highlighting the importance of quality score distribution in MLLM-based IQA methods.

\noindent \textbf{Acknowledgement}.
This work was sponsored by RGC Early Career Scheme (ECS) No. 24209224, National Natural Science Foundation of China (Grant No.62276251), and the Joint Lab of CAS-HK.

{
    \small
    \bibliographystyle{ieeenat_fullname}
    \bibliography{main}
}

\renewcommand\thefigure{A\arabic{figure}}
\renewcommand\thetable{A\arabic{table}}  
\renewcommand\theequation{A\arabic{equation}}
\setcounter{section}{0}
\setcounter{equation}{0}
\setcounter{table}{0}
\setcounter{figure}{0}

\clearpage
\maketitle
\appendix
\section*{Appendix}
\section{Overview}

This \supp is structured as follows.
First, more methodology details are described in \cref{supp:sec:detail}, followed by more ablation studies and qualitative results in \cref{supp:sec:results}.
Then, some extensions are illustrated in \cref{supp:sec:extensions}. 
Finally, we discuss the limitations in \cref{supp:sec:discussion}.

\section{More Details}\label{supp:sec:detail}

\subsection{Details of Dataset Construction}

\textbf{Special designs for PIPAL dataset}.  
The PIPAL dataset is annotated using pair-wise comparisons and Elo ratings, instead of the conventional five-point standard rating. 
As a result, variance information is not provided.  
To integrate the PIPAL dataset into our training framework, we assign a pseudo variance derived from other datasets.  
The statistics of the other three training datasets are shown in \cref{supp:tab:std_statis}.  
Based on these statistics, we manually set the ratio of the mean standard deviation (std) to the score range for the PIPAL dataset as 20\%.  
The minimum and maximum scores in the PIPAL dataset are 934.95 and 1835.99, respectively, thus the pseudo std is set as $20\% \times (1835.99 - 934.95) = 180.21$.  
Note that all scores are normalized to [1,5] during training, with the std / variance normalized accordingly.

\begin{table}[h]
\setlength\tabcolsep{7.8pt}
\centering
\small
\caption{
    \textbf{Statistics} of score range and standard deviation (std). 
}
\vspace{-5pt}
\begin{tabular}{c|cccccc}
\toprule
Datasets & KonIQ & SPAQ & KADID \\
\midrule
score range (max - min) & 2.91 & 91.67 & 3.93 \\
mean std & 0.57 & 13.93 & 0.86 \\
mean std / score range & 19.73\% & 15.20\% & 21.90\% \\
\bottomrule
\end{tabular}
\label{supp:tab:std_statis}
\vspace{-5pt}
\end{table}

\vspace{2pt}\noindent\textbf{Degradation to linear interpolation when the variance is quite small}. 
When the variance is extremely small, integration calculations can introduce errors. 
These errors can lead to strange solutions when directly solving the two constraint conditions for post-adjustment, resulting in $\alpha$ and $\beta$ values that deviate significantly from 1 and 0, respectively. 
Consequently, the adjusted ``probabilities'' can become substantially smaller than 0 or larger than 1, which is unreasonable as the training label. 
Recall that the quality score, $x$, is modeled as a Gaussian distribution, $\mathcal{N}(\mu, \sigma^2)$. 
To address this chanllenge, when the variance is extremely small, we approximate the probability density function, $f(x)$, of the score's Gaussian distribution as a unit impulse function: 
\begin{equation}
    \lim_{\sigma \rightarrow 0} f(x) = \delta(x-\mu), 
\end{equation}
where $\delta(\cdot)$ is the unit impulse function.
In this case, the soft label is calculated through linear interpolation between the two nearest center points. 
Suppose that $c_j < \mu \leq c_{j+1}$, and recall that $d = c_{j+1} - c_j = 1$, the soft label is obtained as: 
\begin{equation}
p_i = 
\left\{
\begin{aligned}
&c_{j+1} - \mu, & \text{if } i = j, \\
&\mu - c_j, & \text{if } i = j+1, \\
&0, & \text{otherwise}.
\end{aligned}
\right.
\end{equation}
According to the $3 \sigma$ rule, nearly all samples fall within the range $[\mu - 3 \sigma, \mu + 3 \sigma]$. 
Thus, if $3 \sigma$ is less than half of the level interval, only the two nearest levels have non-zero probabilities. 
This criterion provides a way to define the threshold for small variance as $3 \sigma \leq d / 2$, which simplifies to $\sigma \leq 0.17$. 
To allow for a slightly relaxed threshold, the threshold for small (normalized) variance is set to $(0.2)^2$.

\subsection{Details of Methodology}

\textbf{Model architecture}. 
\method adopts the architecture of mPLUG-Owl2~\citep{mplugowl2}, structured as follows. 
Specifically, the input images and the question texts are first tokenized, then fused, finally processed by the Large Language Model (LLM) for response generation. 
(a) Tokenizing input images and texts. 
We use a pre-trained CLIP pre-trained ViT-L/14~\citep{clip} as the image encoder to convert the input images into visual tokens, with each token having a channel of 1024. 
The texts are tokenized into textual tokens using the SentencePiece tokenizer~\citep{sentencepiece}, with each token having a channel of 4096. 
To bridge the different embedding spaces of visual and textual tokens, a trainable image abstractor, which is a six-layer transformer network, is implemented to map the vision tokens to the hidden dimension of the LLM, which is 4096. 
The abstractor can also significantly reduce the number of vision tokens to 64, relieving the computing pressure. 
(b) Token fusion. 
We integrate the visual tokens into pre-defined positions within the textual tokens as token fusion. 
(c) Response generation using LLM. 
The fused tokens are fed into an LLM, which is LLaMA-2-7B~\cite{llama2}, to generate the final response. 
The LLM can be either LoRA-tuned~\cite{lora} or fully tuned, and their results are similar when the vision components are trainable, as shown in \cref{supp:tab:ablation_train}.

\begin{figure}[t]
    \centering
    \includegraphics[width=1.0\linewidth]{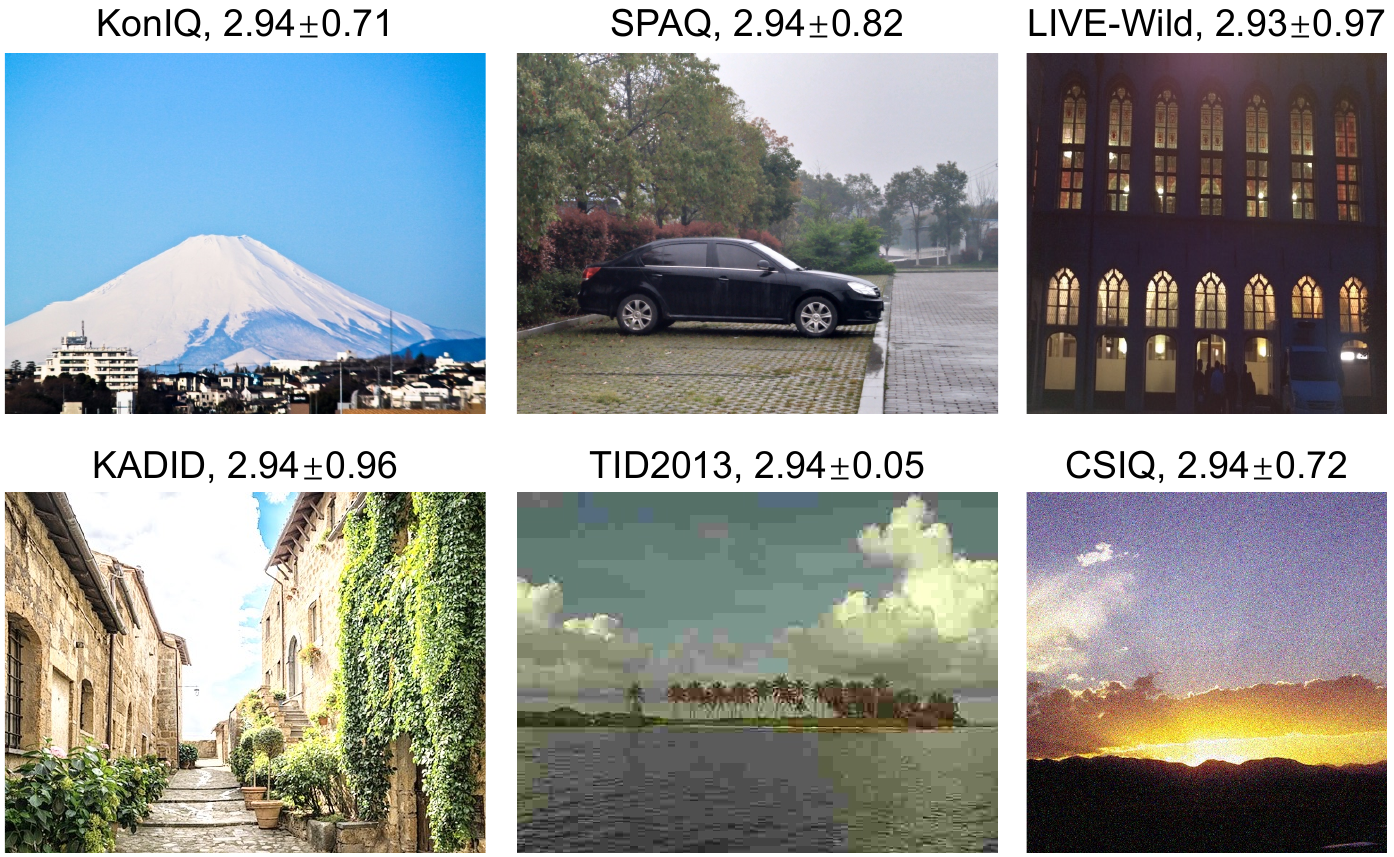}
    \vspace{-15pt}
    \caption{
    \textbf{Examples of dataset variation}. 
    Images from different IQA datasets have similar mean quality scores (\ie, linearly normalized scores), but exhibit significantly different visual quality. 
    The image from KonIQ dataset has the best quality than others. 
    }
    \label{supp:fig:dataset}
    \vspace{-5pt}
\end{figure}

\vspace{2pt}\noindent\textbf{Examples of dataset variation}.  
As discussed in the main paper, different IQA datasets have distinct distributions. 
To illustrate this, we present six images sampled from various IQA datasets in \cref{supp:fig:dataset}.  
Although these images share similar mean quality scores (\ie, linearly normalized scores), they exhibit significantly different visual quality.

\subsection{Details of Training and Inference}

\textbf{Training Setup}. 
The pre-trained weights of mPLUG-Owl2 are used for model initialization. 
The loss weighting term is set to 0.05, bringing the two loss terms to roughly the same scale. 
We adopt AdamW~\cite{adamw} as the optimizer, with an initial learning rate of 2e-5 that gradually decays using a cosine decay strategy. 
A warmup strategy is applied with a warmup ratio of 0.03. 
Our model is trained with a batch size of 64 for 3 epochs.   
Both the vision encoder and abstractor are trainable, and the LLM is fully tuned unless specified. 
As shown in \cref{supp:tab:ablation_train}, LoRA-tuned LLM achieves comparable performance to fully-tuned LLM, providing an alternative for scenarios with limited computation resources. 
Using 8 RTX A6000 GPUs, training on the KonIQ dataset is completed in about 1.5 hours, while training on the KonIQ, SPAQ, and KADID datasets takes around 4 hours.

\vspace{2pt}\noindent\textbf{Explanation of closed-set softmax during inference}. 
In the main paper, we follow Q-Align~\cite{qalign} by applying a closed-set softmax on the five levels to compute the probabilities $p_i^{\text{pred}}$, thereby avoiding the influence of other textual tokens. 
As shown in our statistics in \cref{supp:tab:prob_sum}, the probability sum of the five levels after training is very close to 1, indicating that the trained MLLM consistently predicts one of the five levels. 
Thus, applying softmax over the five levels or across all textual tokens yields nearly identical results.

\begin{table}[t]
\setlength\tabcolsep{2.5pt}
\centering
\footnotesize
\caption{
    \textbf{Probability sum of the five levels} when applying softmax function on all textual tokens. 
}
\vspace{-5pt}
\begin{tabular}{c|cccccc}
\toprule
Datasets & KonIQ & SPAQ & KADID & PIPAL & LIVE-Wild & AGIQA-3K \\
\midrule
Prob. Sum & 0.9998 & 0.9998 & 0.9996 & 0.9997 & 0.9997 & 0.9996 \\
\bottomrule
\end{tabular}
\label{supp:tab:prob_sum}
\vspace{-10pt}
\end{table}

\section{More Results}\label{supp:sec:results}

\begin{figure*}[tb]
    \centering
    \includegraphics[width=1.0\linewidth]{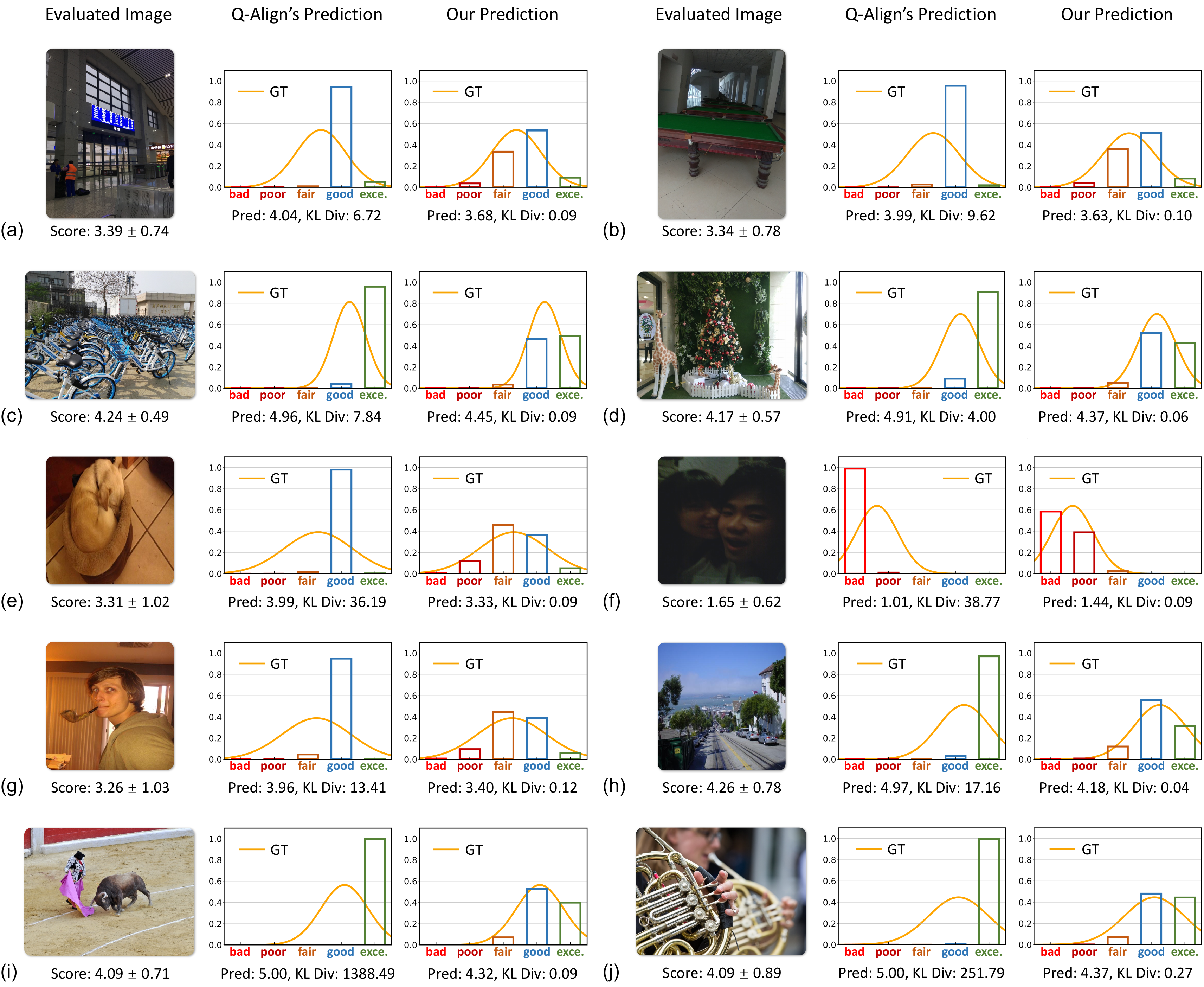}
    \vspace{-10pt}
    \caption{
    \textbf{Qualitative results} on in-the-wild IQA datasets, sampled from the KonIQ, SPAQ, and LIVE-Wild datasets. 
    }
    \label{supp:fig:res_wild}
\end{figure*}

\begin{figure*}[t]
    \centering
    \includegraphics[width=1.0\linewidth]{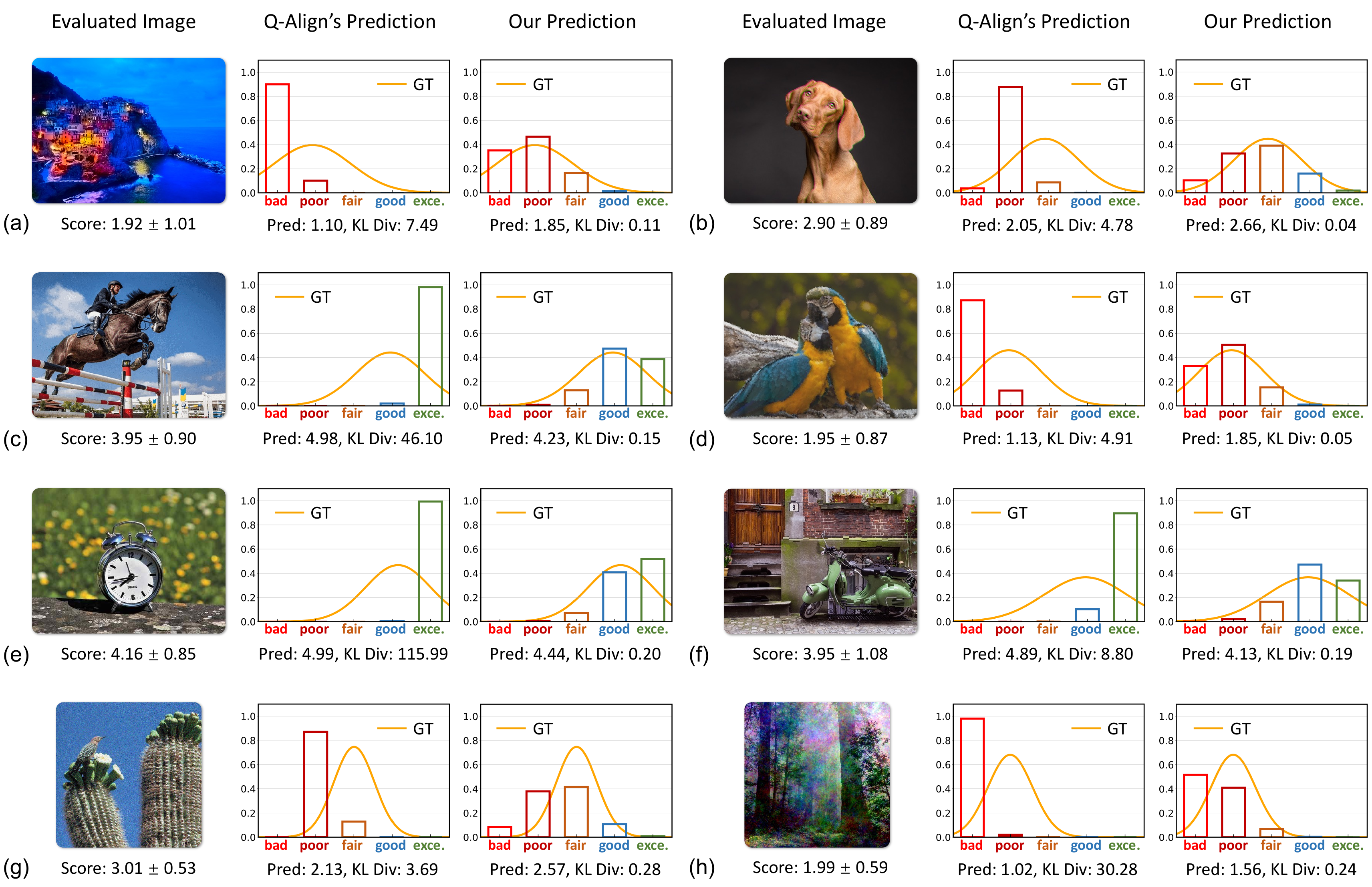}
    \vspace{-10pt}
    \caption{
    \textbf{Qualitative results} on IQA datasets with artificial distortions, sampled from the KADID and CSIQ datasets. 
    }
    \label{supp:fig:res_artificial}
\end{figure*}

\begin{figure*}[t]
    \centering
    \includegraphics[width=1.0\linewidth]{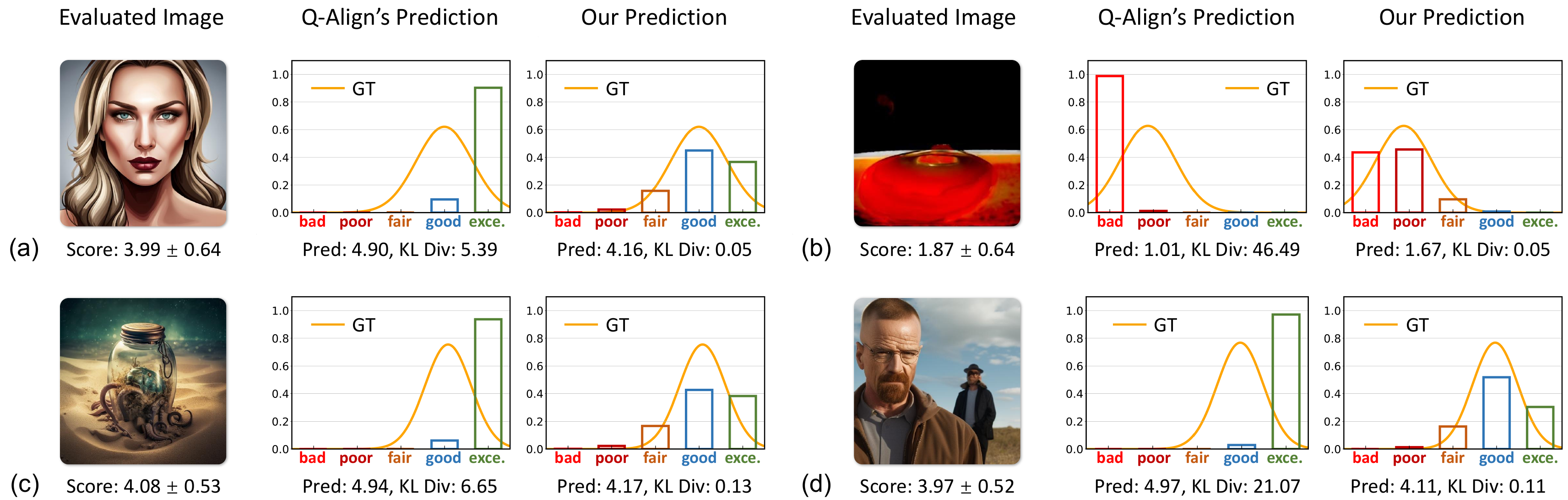}
    \vspace{-10pt}
    \caption{
    \textbf{Qualitative results} on IQA datasets with AI-generated images, sampled from the AGIQA-3K dataset. 
    }
    \label{supp:fig:res_aigc}
\end{figure*}

\textbf{Results of score distribution prediction with KL divergence as metric}. 
In \cref{tab:pred_dist}, we provide the distribution prediction results with JS divergence and Wasserstein distance as metrics. 
Here we further add the KL divergence as metric. 
The KL divergence between two Gaussian distributions, $p_1=\mathcal{N}(\mu_1, \sigma_1^2)$, $p_2=\mathcal{N}(\mu_2, \sigma_2^2)$, is calculated as: 
\begin{equation}
    {\rm KL}(p_1 || p_2) = \log(\frac{\sigma_2}{\sigma_1}) + \frac{\sigma_1^2 + (\mu_1 - \mu_2)^2}{2 \sigma_2^2} - \frac{1}{2}.
    \label{supp:eq:kl}
\end{equation}
The results are given in \cref{supp:tab:pred_dist}, where the KL divergence of Q-Align is quite large. 
We explain this as follows. 
As in \cref{fig:res,supp:fig:res_artificial,supp:fig:res_wild,supp:fig:res_aigc}, Q-Align tends to predict a single level token, bringing a quite small predicted variance. 
That means, the first variance, $\sigma_1^2$ in \cref{supp:eq:kl}, is quite small ($\to 0$), thus the KL divergence goes extremely large ($\to+\infty$).

\begin{table}[t]
\setlength\tabcolsep{3.2pt}
\centering
\footnotesize
\caption{
    \textbf{Score distribution prediction results} with KL divergence between the predicted Gaussian distribution and the human labeled Gaussian distribution as metric. 
    Models are trained on the KonIQ, SPAQ, and KADID datasets. 
    \method achieves a much closer alignment with human annotations. 
}
\vspace{-5pt}
\begin{tabular}{c|ccccc}
\toprule
& KonIQ & SPAQ & KADID & LIVE-Wild & AGIQA-3K \\
\midrule
Q-Align~\cite{qalign} & 109.039 & 2.329 & 1229.530 & 3.980 & 76.414 \\
\method & \textbf{0.058} & \textbf{0.241} & \textbf{0.142} & \textbf{0.249} & \textbf{0.534} \\
\bottomrule
\end{tabular}
\label{supp:tab:pred_dist}
\vspace{-5pt}
\end{table}

\vspace{2pt}\noindent\textbf{Results on more AI-generated images}.
In the main paper, we have included an IQA dataset with AI-generated images, AGIQA-3K~\cite{agiqa}, for evaluation. 
Here we further provide evaluation results on three additional AIGC datasets including AIGCIQA2023~\cite{aigciqa2023}, AGIN~\cite{agin}, and AGIQA-1K~\cite{agiqa1k} in \cref{supp:tab:res_aigc}. 
Both Q-Align and our \method are trained on the KonIQ, SPAQ, and KADID datasets, and then directly evaluated on these unseen AIGC datasets. 
Our \method consistently outperforms the baseline method.

\begin{table}[t]
\centering
\footnotesize
\setlength\tabcolsep{1.8pt}
\caption{
    \textbf{Results on AI-generated images} with PLCC / SRCC metrics. The models are trained on KonIQ, SPAQ, and KADID. 
}
\vspace{-7pt}
\begin{tabular}{c|cccc}
\toprule
& AIGCIQA2023 & AGIN & AGIQA-1K & AGIQA-3K \\
\hline
Q-Align~\cite{qalign} & 
0.809 / 0.783 &
0.655 / 0.615 &
0.650 / 0.442 &
0.788 / 0.733 \\
\method & 
\textbf{0.826} / \textbf{0.799} &
\textbf{0.674} / \textbf{0.626} &
\textbf{0.715} / \textbf{0.514} &
\textbf{0.808} / \textbf{0.745} \\
\bottomrule
\end{tabular}
\label{supp:tab:res_aigc}
\vspace{-10pt}
\end{table}

\begin{table*}[t]
\setlength\tabcolsep{5pt}
\centering
\footnotesize
\caption{
    \textbf{Score regression results of co-training on multiple IQA datasets} with the PLCC / SRCC metrics. 
    The models are trained on the KonIQ, SPAQ, and KADID datasets. 
}
\vspace{-5pt}
\label{supp:tab:more_baseline}
\begin{tabular}{c|ccc|ccccc}
\toprule
Methods & KonIQ & SPAQ & KADID & PIPAL & LIVE-Wild & AGIQA-3K & TID2013 & CSIQ \\
\midrule
LIQE~\cite{liqe} & 
0.907 / 0.922 &
0.916 / 0.921 &
0.928 / 0.929 &
\textbf{0.503} / 0.493 &
0.855 / 0.822 &
0.689 / 0.650 &
\textbf{0.881} / \textbf{0.854} &
0.792 / 0.794 \\
Compare2Score~\cite{compare2score} & 
0.941 / 0.929 &
0.929 / 0.927 &
0.952 / 0.949 &
0.446 / 0.440 &
0.868 / 0.856 &
0.787 / 0.733 &
0.836 / 0.809 &
0.879 / 0.830 \\
Q-Align~\cite{qalign} & 
0.945 / 0.938 & 
0.933 / 0.931 & 
0.935 / 0.934 & 
0.409 / 0.420 & 
0.887 / 0.883 & 
0.788 / 0.733 & 
0.829 / 0.808 & 
0.876 / 0.845 \\
\method & 
\textbf{0.957} / \textbf{0.944} & 
\textbf{0.938} / \textbf{0.934} & 
\textbf{0.955} / \textbf{0.953} & 
0.495 / \textbf{0.496} & 
\textbf{0.900} / \textbf{0.887} & 
\textbf{0.808} / \textbf{0.745} & 
0.852 / 0.820 & 
\textbf{0.900} / \textbf{0.857} \\
\bottomrule
\end{tabular}
\vspace{-5pt}
\end{table*}

\begin{table*}[t]
\setlength\tabcolsep{6.2pt}
\centering
\footnotesize
\caption{
    \textbf{Ablation studies on level numbers} with PLCC / SRCC metrics. We use numerical names (one / two / three / four / five / ...) as level names because they are easy to extend to different numbers. The models are trained on the KonIQ, SPAQ, and KADID datasets. 
}
\vspace{-5pt}
\label{supp:tab:ablation_level_number}
\begin{tabular}{c|ccc|ccccc}
\toprule
Level Number & KonIQ & SPAQ & KADID & PIPAL & LIVE-Wild & AGIQA-3K & TID2013 & CSIQ \\
\midrule
5 &
0.945 / 0.932 &
0.934 / 0.930 &
0.955 / 0.952 &
\textbf{0.421} / \textbf{0.424} &
0.873 / 0.857 &
0.766 / 0.712 &
\textbf{0.820} / \textbf{0.794} &
\textbf{0.882} / \textbf{0.835} \\
6 & 
0.947 / 0.935 &
0.935 / 0.931 &
\textbf{0.956} / \textbf{0.953} &
0.404 / 0.408 &
0.879 / 0.867 &
\textbf{0.789} / \textbf{0.732} &
0.808 / 0.789 &
0.877 / 0.830 \\
7 &
0.947 / 0.935 &
0.935 / \textbf{0.932} &
0.952 / 0.949 &
0.411 / 0.413 &
\textbf{0.883} / \textbf{0.869} &
0.777 / 0.720 &
0.812 / 0.789 &
0.879 / 0.827 \\
8 & 
\textbf{0.949} / \textbf{0.936} &
\textbf{0.936} / \textbf{0.932} &
0.949 / 0.946 &
0.420 / 0.419 &
0.872 / 0.854 &
0.772 / 0.709 &
0.795 / 0.779 &
0.874 / 0.824 \\
10 & 
0.932 / 0.916 &
0.930 / 0.927 &
0.948 / 0.942 &
0.409 / 0.403 &
0.861 / 0.837 &
0.737 / 0.654 &
0.784 / 0.763 &
0.880 / 0.831 \\
12 & 
0.930 / 0.916 &
0.930 / 0.927 &
0.947 / 0.943 &
0.415 / 0.412 &
0.864 / 0.840 &
0.722 / 0.643 &
0.804 / 0.773 &
0.870 / 0.819 \\
\bottomrule
\end{tabular}
\vspace{7pt}
\setlength\tabcolsep{2pt}
\centering
\footnotesize
\caption{
    \textbf{Ablation studies on training / fixing various model components} with PLCC / SRCC metrics. 
    ``Enc.'' means vision encoder, and ``Abs.'' represents vision abstractor. ``LLM LoRA'' or ``LLM Full'' represents the LLM is LoRA-tuned~\cite{lora} or fully tuned. ``\ding{52}'' means this component is trained. 
    The models are trained on the KonIQ, SPAQ, and KADID datasets. 
    The results show that training vision encoder and abstractor significantly improves the performance. 
    When vision encoder and abstractor are trained, fully-tuned LLM only shows a slight advantage over LoRA-tuned LLM. 
}
\vspace{-5pt}
\label{supp:tab:ablation_train}
\begin{tabular}{c|cccc|ccc|ccccc}
\toprule
\# & Enc. & Abs. & LLM LoRA & LLM Full & KonIQ & SPAQ & KADID & PIPAL & LIVE-Wild & AGIQA-3K & TID2013 & CSIQ \\
\midrule
0 & & & \ding{52} & & 
0.826 / 0.797 &
0.871 / 0.867 &
0.814 / 0.803 &
0.440 / 0.427 &
0.728 / 0.695 &
0.804 / 0.750 &
0.740 / 0.692 &
0.818 / 0.751 \\
1 & & \ding{52} & \ding{52} & & 
0.909 / 0.890 &
0.919 / 0.916 &
0.898 / 0.892 &
0.429 / 0.425 &
0.806 / 0.768 &
0.699 / 0.672 &
0.749 / 0.682 &
0.876 / 0.821 \\
2 & \ding{52} & \ding{52} & \ding{52} & &
0.955 / 0.942 &
\textbf{0.938} / \textbf{0.934} &
0.953 / 0.950 &
0.479 / 0.473 &
0.898 / 0.884 &
\textbf{0.810} / \textbf{0.756} &
0.849 / \textbf{0.824} &
\textbf{0.900} / \textbf{0.861} \\
3 & & & & \ding{52} &
0.889 / 0.867 &
0.913 / 0.910 &
0.873 / 0.866 &
0.429 / 0.417 &
0.784 / 0.747 &
0.762 / 0.711 &
0.743 / 0.670 &
0.843 / 0.759 \\
4 & & \ding{52} & & \ding{52} & 
0.911 / 0.893 &
0.920 / 0.916 &
0.905 / 0.899 &
0.439 / 0.429 &
0.819 / 0.781 &
0.764 / 0.705 &
0.765 / 0.699 &
0.878 / 0.820 \\
5 & \ding{52} & \ding{52} & & \ding{52} & 
\textbf{0.957} / \textbf{0.944} &
\textbf{0.938} / \textbf{0.934} &
\textbf{0.955} / \textbf{0.953} &
\textbf{0.495} / \textbf{0.496} &
\textbf{0.900} / \textbf{0.887} &
0.808 / 0.745 &
\textbf{0.852} / 0.820 &
\textbf{0.900} / 0.857 \\
\bottomrule
\end{tabular}
\vspace{-5pt}
\end{table*}

\begin{table*}[t]
\setlength\tabcolsep{8.5pt}
\centering
\footnotesize
\caption{
    \textbf{Low-level perception results on Q-Bench~\cite{qbench}} of co-training with or pre-training on score regression tasks. 
    The results of \#0, \#1, and \#2 are borrowed from~\cite{qinstruct}. 
    Co-training with or pre-training on score regression tasks stably improves the performance. 
}
\vspace{-5pt}
\begin{tabular}{c|c|ccccccc|c}
\toprule
\# & Method & Yes / No & What & How & Distortion & Other & IC Distortion & IC Other & Overall \\
\midrule
0 & From the scratch~\cite{qinstruct} & 0.7218 & 0.5796 & 0.5619 & 0.5668 & 0.6921 & 0.5329 & 0.7265 & 0.6161 \\
1 & High-level co-training~\cite{qinstruct} & 0.7564 & 0.6704 & 0.5903 & 0.7101 & 0.6528 & 0.6316 & 0.6980 & 0.6756 \\
2 & High-level pre-training~\cite{qinstruct} & 0.7600 & 0.6504 & 0.6166 & 0.6595 & 0.6875 & 0.6546 & 0.7388 & 0.6796 \\
\midrule
3 & Score co-training & 0.7727 & 0.6615 & 0.6308 & 0.6965 & 0.6875 & 0.6349 & 0.7633 & 0.6925 \\
4 & Score pre-training & \textbf{0.7927} & \textbf{0.7323} & \textbf{0.6410} & \textbf{0.7276} & \textbf{0.7060} & \textbf{0.6974} & \textbf{0.7837} & \textbf{0.7258} \\
\bottomrule
\end{tabular}
\label{supp:tab:description}
%
\setlength\tabcolsep{3.5pt}
\centering
\footnotesize
\vspace{8pt}
\caption{
    \textbf{Co-training with Q-Instruct~\cite{qinstruct} leads to an obvious reduction in score regression performance}. 
    Exploring better strategies to combine language-based quality description tasks with score regression tasks is left as our future work. 
}
\vspace{-5pt}
\label{supp:tab:joint}
\begin{tabular}{c|ccc|ccccc}
\toprule
& KonIQ & SPAQ & KADID & PIPAL & LIVE-Wild & AGIQA-3K & TID2013 & CSIQ \\
\midrule
Co-training with Q-Instruct & 
0.916 / 0.926 & 
0.870 / 0.816 & 
0.812 / 0.810 & 
0.286 / 0.289 & 
0.825 / 0.812 & 
0.778 / 0.732 & 
0.680 / 0.656 & 
0.798 / 0.790 \\
Only score regression & 
\textbf{0.957} / \textbf{0.944} &
\textbf{0.938} / \textbf{0.934} &
\textbf{0.955} / \textbf{0.953} &
\textbf{0.495} / \textbf{0.496} &
\textbf{0.900} / \textbf{0.887} &
\textbf{0.808} / \textbf{0.745} &
\textbf{0.852} / \textbf{0.820} &
\textbf{0.900} / \textbf{0.857} \\
\bottomrule
\end{tabular}
\vspace{-5pt}
\end{table*}

\vspace{2pt}\noindent\textbf{More baselines on multi-dataset training}. 
In the main paper, we primarily compare with Q-Align. 
Here we add the comparison results with more baselines including LIQE~\cite{liqe} and Compare2Score~\cite{compare2score} in \cref{supp:tab:more_baseline}. 
Our \method consistently outperforms these baselines.

\vspace{2pt}\noindent\textbf{Ablation studies on level number} are conducted in \cref{supp:tab:ablation_level_number}. 
The performance improves with moderately larger level numbers (\ie, 6$\sim$8), as discretization becomes more accurate. 
However, the performance declines with much larger level numbers (\ie, 10$\sim$12), due to the increased difficulty in level prediction (more classification categories).

\vspace{2pt}\noindent\textbf{Ablation studies on training / fixing components} are summarized in \cref{supp:tab:ablation_train}.  
The vision encoder and abstractor can be either fixed or trained, while the LLM can be LoRA-tuned~\cite{lora} or fully tuned.  
First, comparing \#0 with \#1, or \#3 with \#4, training the vision abstractor significantly enhances performance.  
Second, similarly, from \#1 to \#2, or \#4 to \#5, training the vision encoder also leads to performance improvements.  
This improvement may be attributed to the fact that training either the vision encoder or abstractor helps extract more relevant features for IQA.  
Third, comparing \#0 with \#3, when the vision components are fixed, fully-tuned LLM shows a substantial advantage over LoRA-tuned LLM.  
Finally, from \#1 to \#4, or \#2 to \#5, when the vision components are trainable, fully-tuned LLM demonstrates similar performance to LoRA-tuned LLM.  
This provides an alternative if the computation resources are limited.

\vspace{2pt}\noindent\textbf{Qualitative results} are provided in \cref{supp:fig:res_artificial,supp:fig:res_wild,supp:fig:res_aigc}. 
Q-Align, which is trained on one-hot labels, tends to predict a single label, resulting in higher KL divergence. 
Our \method can predict the score distribution that aligns well with human annotations under wide circumstances: 
\begin{itemize}
    \item Different categories of distortions, including \textit{in-the-wild} images with authentic distortions in \cref{supp:fig:res_wild}, \textit{artificial} distortions in \cref{supp:fig:res_artificial}, \textit{AI-generated} images in \cref{supp:fig:res_aigc}. 
    \item Multiple quality levels, such as, \textit{poor} quality (\cref{supp:fig:res_wild}f, \cref{supp:fig:res_artificial}ah), \textit{fair} quality (\cref{supp:fig:res_wild}e, \cref{supp:fig:res_artificial}g), and \textit{high} quality (\cref{supp:fig:res_wild}hj, \cref{supp:fig:res_aigc}c). 
    \item Various image contents, including \textit{animals} (\cref{supp:fig:res_wild}e, \cref{supp:fig:res_artificial}bd), \textit{humans} (\cref{supp:fig:res_wild}fg, \cref{supp:fig:res_aigc}ad), \textit{nature scenes} (\cref{supp:fig:res_artificial}agh), \textit{urban scenes} (\cref{supp:fig:res_wild}ch, \cref{supp:fig:res_artificial}f), \textit{indoor scenes} (\cref{supp:fig:res_wild}abd), and \textit{sports} (\cref{supp:fig:res_wild}i, \cref{supp:fig:res_artificial}c). 
\end{itemize}

\section{Extensions}\label{supp:sec:extensions}

\textbf{Score regression helps quality description}. 
We investigate whether score regression tasks can enhance general low-level perception tasks, \ie, language-based quality description tasks. 
The low-level perception tasks are evaluated on Q-Bench~\cite{qbench}. 
For score regression, we adopt the same training techniques as in the main paper, and train the model on the KonIQ, SPAQ, and KADID datasets. 
The low-level perception tasks are trained using the next token prediction paradigm, with Q-Instruct~\cite{qinstruct} as the training dataset. 
Q-Instruct~\cite{qinstruct} has shown that co-training with or pre-training on high-level tasks can improve performance on low-level perception tasks. 
Similarly, we evaluate two strategies: 
(a) co-training with score regression tasks, and 
(b) pre-training on score regression tasks. 
Some questions in the Q-Instruct dataset are quite similar to the questions of score regression, which may confuse the model. 
Therefore, we append the questions of score regression with a specific suffix, ``Answer the question with levels.'', to specify the task.

The experimental results are presented in~\cref{supp:tab:description}. 
First, comparing \#0 with \#3 \& \#4, both co-training with and pre-training on score regression tasks consistently improve low-level perception performance. 
Second, comparing \#1 with \#3, or \#2 with \#4, the benefits from score regression tasks are greater than those from high-level tasks, likely because score regression is more closely related to low-level perception. 
Finally, from \#3 to \#4, pre-training on score regression tasks achieves better results than co-training.

\vspace{2pt}\noindent\textbf{Quality description harms score regression}. 
Considering the promising results in \cref{supp:tab:description}, we aim to explore whether it is possible to co-train a model for both accurate score regression and language-based quality description. 
We therefore evaluate the score regression results of the co-trained model in \cref{supp:tab:joint}. 
These preliminary results indicate that co-training with the instruction-tuning dataset, Q-Instruct, leads to a noticeable decrease in score regression. 
The reason can be that, in quality description datasets, the words to describe the quality levels are very diverse, greatly beyond the pre-defined five levels for score regression. 
This may confuse the model when predicting the level tokens.

\begin{figure}[t]
    \centering
    \includegraphics[width=1.0\linewidth]{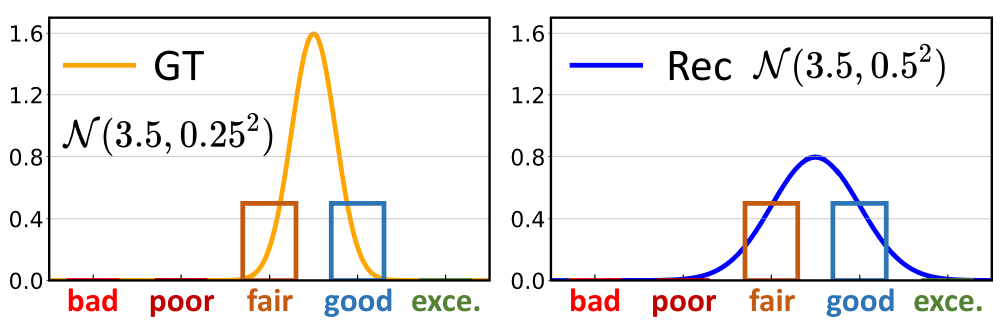}
    \vspace{-15pt}
    \caption{
    \textbf{Illustration of discretization errors when the variance is quite small}. 
    A score distribution, $\mathcal{N}(3.5, 0.25^2)$, is discretized into a soft label, where both ``fair'' and ``good'' share a probability of 0.5. 
    However, the recovered distribution from this soft label becomes $\mathcal{N}(3.5, 0.5^2)$, resulting in a larger variance and, consequently, a flatter curve of the distribution density function. 
    }
    \label{supp:fig:error}
    \vspace{-10pt}
\end{figure}

\vspace{-3pt}
\section{Limitations and Future Works}\label{supp:sec:discussion}
\vspace{-2pt}

\textbf{Simple co-training with quality description cannot improve score regression}. 
As shown in \cref{supp:sec:extensions}, score regression tasks can enhance quality description results, while co-training with quality description tasks reduces the score regression performance. 
How to better co-train these two tasks remains an open question. 
However, first, the performance of the co-training model is still reasonable and higher than many non-MLLM-based IQA methods. 
Second, with the rapid development of MLLM-based quality description research, better quality description datasets may help.

\vspace{2pt}\noindent\textbf{Our discretization introduces errors when the variance is very small}. 
While our discretization method effectively handles most distributions, it still introduces errors in the variance of the distribution when the variance is very small. 
For instance, as shown in \cref{supp:fig:error}, a score distribution with a small variance, $\mathcal{N}(3.5, 0.25^2)$, is discretized into a soft label where both ``fair'' and ``good'' share a probability of 0.5. 
The recovered distribution from this soft label becomes $\mathcal{N}(3.5, 0.5^2)$, resulting in a larger variance and a flatter curve in the probability density function. 
However, samples with extremely small variance are rare, 
\eg, only 0.29\% of samples in the KonIQ dataset have a variance smaller than $0.5^2$. 
Thus, the overall influence of this issue is relatively small. 
How to better preserve the distribution characteristics during discretization for such cases is our future work.

\vspace{2pt}\noindent\textbf{High memory consumption}.  
As our \method is based on an MLLM with 7B parameters, it requires 15.2G storage space (in bfloat16 format) and 15.8G (batch size 1) / 23.0G (batch size 64) CUDA memory for inference. 
Despite this, it remains deployable on consumer GPUs like 4090, and the memory consumption can be reduced through quantization.

\end{document}